\documentclass[10pt,twocolumn,letterpaper]{article}

\usepackage{cvpr}              

\newcommand{\ours}{ORCA\xspace}
\usepackage{lipsum}
\usepackage{graphicx}
\usepackage{multirow}
\usepackage{amsmath}
\usepackage{amssymb}
\usepackage{multicol}
\usepackage{multirow}
\usepackage[accsupp]{axessibility}
\usepackage[ruled,vlined]{algorithm2e}
\newcommand{\PyCode}[1]{\ttfamily\textcolor{black}{#1}} 

\definecolor{cvprblue}{rgb}{0.21,0.49,0.74}
\usepackage[pagebackref,breaklinks,colorlinks,allcolors=cvprblue]{hyperref}

\def\paperID{*****} 
\def\confName{CVPR}
\def\confYear{2026}

\title{Exploring Conditions for Diffusion Models in Robotic Control}

\author{Heeseong Shin$^{1*}$ \quad Byeongho Heo$^{2}$ \quad Dongyoon Han$^{2}$ \quad Seungryong Kim$^{1\dagger}$
\quad Taekyung Kim$^{2\dagger}$\\
\\
\textsuperscript{1}KAIST AI \qquad \textsuperscript{2}NAVER AI Lab\\
{\url{https://orca-rc.github.io/}}
}

\begin{document}

\twocolumn[{
\renewcommand\twocolumn[1][]{#1}%
\maketitle
\begin{center}
  \vspace{-10pt}
  \includegraphics[width=1.0\linewidth]{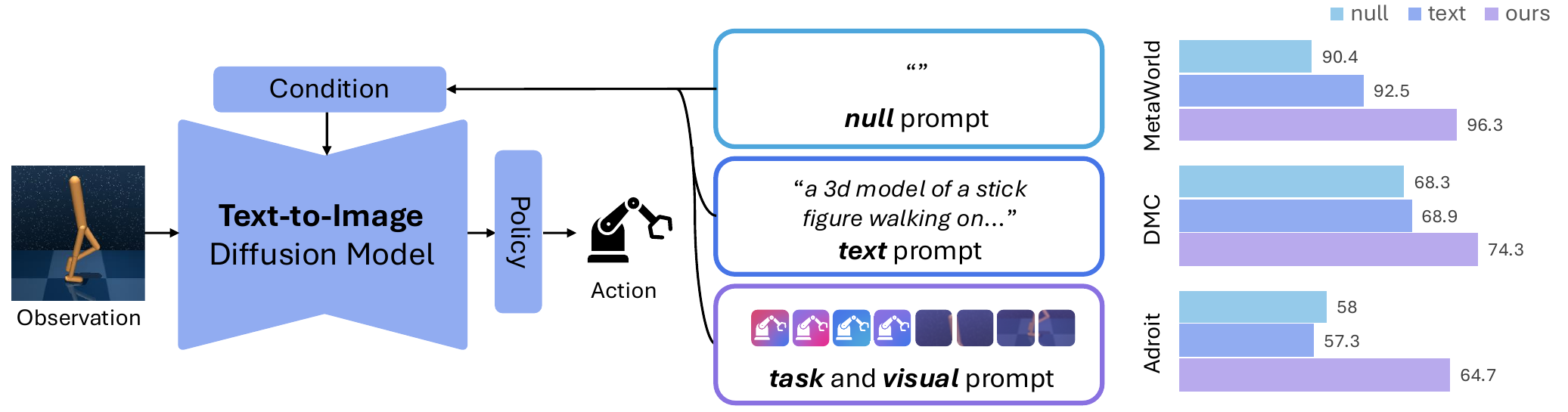}
  \captionof{figure}{\textbf{How can we condition diffusion models for robotic control?} We investigate methods for \textit{conditioning} text-to-image diffusion models~\citep{rombach2022high} to perform robotic control, aiming to address various tasks in a \textit{task-adaptive} manner. We observe that text prompts, unlike in other vision tasks~\citep{zhao2023unleashing}, are ineffective for robotic control. Therefore, we propose to learn \textbf{task} prompts in control environments and further incorporate dynamic details through \textbf{visual} prompts for conditioning diffusion models.}
  \label{fig:overview}
\end{center}
}]

\maketitle{}

\let\thefootnote\relax\footnotetext{$^*$Work done during an internship at NAVER AI Lab. 
\\
\phantom{000.}$^\dagger$ Corresponding authors.}

\begin{abstract}
While pre-trained visual representations have significantly advanced imitation learning, they are often task-agnostic as they remain frozen during policy learning. In this work, we explore conditions for pre-trained text-to-image diffusion models to obtain task-adaptive visual representations for robotic control, without fine-tuning the model itself. We find that naively applying textual conditions—a successful strategy in other vision domains—yields minimal or even negative gains in control tasks. We attribute this to the domain gap between the diffusion model's training data and robotic control environments, leading us to argue for conditions that consider the specific visual information required for control. To this end, we propose \textbf{\ours} for leveraging diffusion models with c\textbf{\underline{o}}nditions in \textbf{\underline{r}}obotic \textbf{\underline{c}}ontrol in a task-\textbf{\underline{a}}daptive manner, which introduces learnable \textit{task} prompts that adapt to the control environment and \textit{visual} prompts that capture fine-grained, frame-specific details. Through facilitating task-adaptive representation, \ours achieves state-of-the-art on various robotic control benchmarks. 
\end{abstract}
    
\section{Introduction}

Recent advances in diffusion models~\citep{ho2020denoising, rombach2022high} have not only facilitated high-quality image synthesis, but also demonstrated as a strong visual representation for various vision tasks~\citep{baranchuk2021label, tang2023emergent, xiang2023denoising}. Among them, pre-trained text-to-image diffusion models, \textit{e.g.} Stable Diffusion~\citep{rombach2022high}, have shown that utilizing text \textit{conditions} can significantly boost performance in visual perception tasks, without the need for fine-tuning the model~\citep{zhao2023unleashing}. The key to leveraging text conditions lies in obtaining well-designed prompts~\citep{kondapaneni2024text}—often describing objects in the image or the given task—that can funnel useful information into downstream tasks. This not only enhances the proficiency of diffusion models on downstream tasks but also broadens their applicability to a wider variety of vision tasks~\citep{yin2025knowledge, wu2025textsplat}.

Robotic control, meanwhile, has also benefited greatly from the introduction of pre-trained visual representations to imitation learning~\citep{parisi2022unsurprising}. By leveraging frozen visual encoders pre-trained on large-scale data, these representations have replaced the previous \textit{tabula-rasa} paradigm of training vision encoders from scratch on limited-scale control data.  However, this approach is limited by its \textbf{task-agnostic} nature, as the visual representations remain frozen during downstream policy learning. Since the suitability of a representation for a specific task is unknown beforehand, determining which representation performs best often requires manual, task-by-task inspection~\citep{majumdar2023we}, which becomes cumbersome given the vast variety of control tasks. While a straightforward solution might be to fine-tune the vision encoder, this often results in poor results as the model loses generalization capabilities by overfitting to specific scenes in imitation learning~\citep{parisi2022unsurprising, hansen2022pre,majumdar2023we}.

In this work, we explore bridging pre-trained text-to-image diffusion models to robotic control for achieving \textbf{task-adaptive} visual representations through \textbf{conditions}, without fine-tuning the diffusion model. Inspired by the effectiveness of conditions in visual perception tasks, we ask the following question: \textit{How can we effectively implement conditions for diffusion models in robotic control?} We begin by investigating textual conditions, generating captions with a state-of-the-art vision-language model~\citep{comanici2025gemini} to observe their impact on control task performance. However, as shown in Fig.~\ref{fig:overview}, the gains are minimal, and in some cases, performance even declines. This result contrasts sharply with findings in other vision tasks~\citep{zhao2023unleashing}, where machine-generated captions have served as strong conditions~\citep{kondapaneni2024text}. 

Upon investigation, we find that pre-trained diffusion models often struggle to accurately associate text conditions to the image in control environments. We attribute this discrepancy to the nature of the diffusion model being trained on web-collected images, which suits visual tasks that involve real-world images and common objects, such as semantic segmentation~\citep{zhao2023unleashing,kondapaneni2024text}. However, control environments, featuring specialized robotic agents performing specific tasks, would require a more careful and deliberate approach to devising effective conditions for downstream policy learning.

Robot control further complicates this challenge, as these tasks operate on dynamic video streams and require a finer visual granularity to interact with specific parts of objects, not just to categorize them. This dynamic nature implies that effective conditions must be generated uniquely for each frame~\citep{hong2024large} to guide evolving actions and adapt to changing visual states. Consequently, we hypothesize that conditions for control should incorporate visual information from every frame to capture both dynamic behavior and fine-grained details. 

These observations suggest that conditions for control should be task-grounded and frame-sensitive. 
To this end, we propose a simple, yet effective method that incorporates visual information while addressing the limitations of text conditions. Specifically, we replace the text prompt with learnable \textbf{task} prompts, which are learned during downstream control tasks to ensure accurate grounding within the specific environment. Furthermore, to enable the conditions to capture the detailed visual state of each frame, we employ a vision encoder and utilize its representations as \textbf{visual} prompts. We demonstrate that both the task and visual prompts can be learned end-to-end during downstream policy learning using a standard behavior cloning objective.

Our framework for leveraging diffusion models with c\textbf{\underline{o}}nditions in \textbf{\underline{r}}obotic \textbf{\underline{c}}ontrol in a task-\textbf{\underline{a}}daptive manner, \textbf{\ours}, achieves state-of-the-art performance in robotic control tasks~\citep{tassa2018deepmind, metaworld, adroit}, surpassing VC-1~\citep{majumdar2023we}. We verify our design choices by comparing to baselines with text conditions and different conditioning methods~\citep{zhou2022learning,kondapaneni2024text} from visual perception tasks. In addition, we provide detailed analysis and ablations on our approach, highlighting the importance of conditions in diffusion models for robotic control.

\section{Related Work}

\paragraph{Pre-trained visual representations for robotic control.}
In recent years, visual representations derived from self-supervised pre-trained models~\citep{radford2021learning, cherti2023reproducible, majumdar2023we, he2022masked, caron2021emerging, kim2025token, kim2023lut, kim2025morphing} have demonstrated notable effectiveness in visuo-motor manipulation tasks~\citep{parisi2022unsurprising}.  Specifically, Parisi et al.~\citep{parisi2022unsurprising} showed that visual representations from frozen pre-trained encoders, such as MoCo~\citep{he2020momentum} and CLIP~\citep{radford2021learning}, can not only outperform representations trained from scratch but are also comparable to ground-truth state features in behavior cloning. 
This finding has motivated subsequent work on pre-trained visual representations for robotic control.
Among these, R3M~\citep{nair2022r3m} employs a time-contrastive learning objective on ego-centric data with vision-language alignment, whereas VIP~\citep{ma2022vip} introduces value-implicit learning to associate goal and initial states. MVP~\citep{radosavovic2023robotmae} and VC-1~\citep{majumdar2023we} both adopt MAE~\citep{he2022masked} pre-training methodologies, curating large datasets that include ego-centric and instructional videos to enhance transferability to robotic manipulation tasks. More recently, SCR~\citep{gupta2024pre} has investigated representations from Stable Diffusion~\citep{rombach2022high} for navigation and control tasks. Nonetheless, these methods opted for keeping the visual representation frozen, resulting them to be task-agnostic. 

\begin{figure*}[t]
  \centering
  \includegraphics[width=\linewidth]{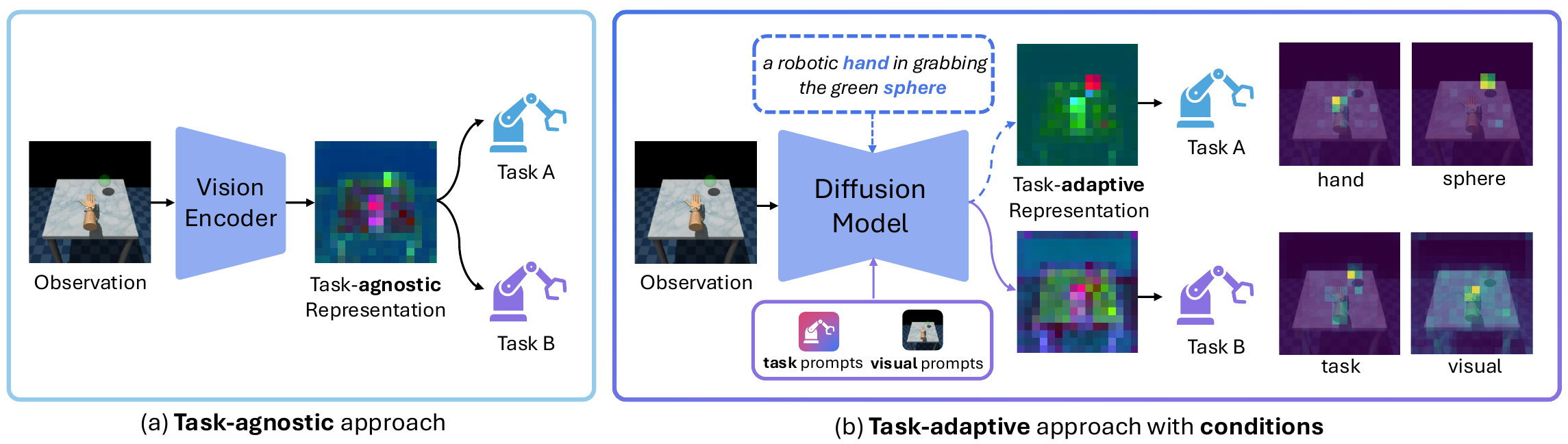}
  \vspace{-15pt}
  \caption{\textbf{Motivation.} We aim to overcome the limitations of existing \textbf{task-agnostic} approach (a) with frozen pre-trained visual representations~\citep{parisi2022unsurprising}, by leveraging \textbf{conditions} in diffusion models for robotic control tasks in a \textbf{task-adaptive} approach (b). In this regard, we explore text conditions(\textbf{\S~\ref{text}}), more advanced methods(\textbf{\S~\ref{discussion},\S~\ref{method}}) as conditions.}
  \label{fig:motivation}
  \vspace{-10pt}
\end{figure*}

\paragraph{Diffusion models as pre-trained visual representations.}
Recent advancements in diffusion models~\citep{ho2020denoising, rombach2022high} have enabled the synthesis of high-resolution images with unprecedented fidelity. This progress has concurrently motivated diverse investigations into the internal representations of generative diffusion models~\citep{tang2023emergent, luo2023diffusion, baranchuk2021label, zhao2023unleashing, xiang2023denoising, wu2025textsplat, kim2025seg4diff} for various downstream vision tasks. This allowed diffusion models to outperform prior approaches with self-supervised pre-trained models~\citep{shin2024towards, hong2022cost} in tasks such as semantic correspondence~\citep{cho2021cats}, semantic segmentation~\citep{Cho_2024_CVPR}, and even 3D reconstruction~\citep{hong2024pf3plat, hong2024unifying}. DDPMSeg~\citep{baranchuk2021label} was among the first to explore the efficacy of diffusion model's representations in label-scarce segmentation, while DDAE~\citep{xiang2023denoising} focused on image classification. DIFT~\citep{tang2023emergent}, DHF~\citep{luo2023diffusion} and SD-DINO~\citep{zhang2023tale} have demonstrated that the representation from diffusion models can achieve state-of-the-art in semantic correspondence tasks. Notably, VPD~\citep{zhao2023unleashing} demonstrated that downstream performance can be enhanced by with text conditions, such as the names of objects present in an image, in tasks such as semantic segmentation and monocular depth estimation. SD4Match~\citep{li2024sd4match} and EcoDepth~\citep{patni2024ecodepth} proposed prompting modules to derive conditions for semantic correspondence and monocular depth estimation. TADP~\citep{kondapaneni2024text} demonstrated that text descriptions generated from vision-language models can serve as strong conditions, and could be further enhanced with style modifiers learned from Textual Inversion~\citep{gal2022image}. However, we distinguish our approach by focusing on robotic control, rather than for visual tasks in general image domains.
\section{Preliminaries}

\textbf{Diffusion models}~\citep{sohl2015deep, ho2020denoising, kingma2021variational} constitute a class of generative models that learn to reverse a multi-step noising process, thereby reconstructing a target data distribution. In this work, we focus on conditional diffusion models (\textit{e.g.} Stable Diffusion~\citep{rombach2022high}), which enable image generation guided by a condition  $\mathcal{C}$, often being text prompts. The training objective is to reverse the noising process, typically discretized into $T$ timesteps. A pre-defined noise schedule, denoted by $\alpha_t$, facilitates the definition of the noised latent variable $z_t$ at timestep $t$ as:
\begin{equation}
    z_t = \sqrt{\bar{\alpha}_t}z_0 + \sqrt{1-\bar{\alpha}_t}\epsilon,
    \label{eq:eq1}
\end{equation}
where $z_0$ is the initial clean data, $\bar{\alpha}_t=\prod^t_{i=1}\alpha_i$, 
and $\epsilon \sim \mathcal{N}(0, I)$ is Gaussian noise. Following Ho et al.~\citep{ho2020denoising}, with appropriate parameterization, diffusion models can be trained by regressing the added noise 
$\epsilon$ from $z_t$:
\begin{equation}
    \mathcal{L}_{\text{DM}} = \mathbb{E}_{z_0, \epsilon, t} \left[ \left\| \epsilon - \epsilon_{\theta}(z_t(z_0, \epsilon), t; \mathcal{C}) \right\|_2^2 \right],
    \label{eq:eq2}
\end{equation}
where $\epsilon_\theta$ indicates the denoising network, typically a U-Net~\citep{ronneberger2015u} or a Transformer~\citep{vaswani2017attention} architecture. 
Stable Diffusion, for our case, is a Latent Diffusion Model (LDM)~\citep{rombach2022high} with an U-Net architecture, in which the diffusion process occurs in a compressed latent space learned by an autoencoder, specifically a VQGAN~\citep{esser2021taming}. For conditional generation, U-Net-based LDMs implement Transformer blocks with cross-attention layers into the U-Net blocks to inject the condition $\mathcal{C}$ into the image generation process.

\paragraph{Extracting visual representation from diffusion models.}
To extract visual representations, initially, an input image $I$ is encoded into its latent representation $z_0=\mathcal{E}(I)$ using the VQGAN encoder $\mathcal{E}$. For a chosen fixed timestep $t$, the corresponding noisy latent $z_t$ is computed via Eq.~\ref{eq:eq1}. This $z_t$ is then processed by the denoising U-Net 
$\epsilon_\theta(\cdot)$. However, as the network $\epsilon_{\theta}$ is trained to predict noise as shown in Eq.~\ref{eq:eq2}, we instead extract intermediate feature maps from within the U-Net~\citep{meng2024not}. We denote the set of extracted intermediate features as $f$, and denote $f=\epsilon_\theta(z_t, t; \mathcal{C})$ to be the output of $\epsilon_\theta$ for simplicity, and primarily consider features from the earlier blocks of the U-Net.

\section{Motivation}
\begin{figure*}[t]
  \centering
  \includegraphics[width=1\linewidth]{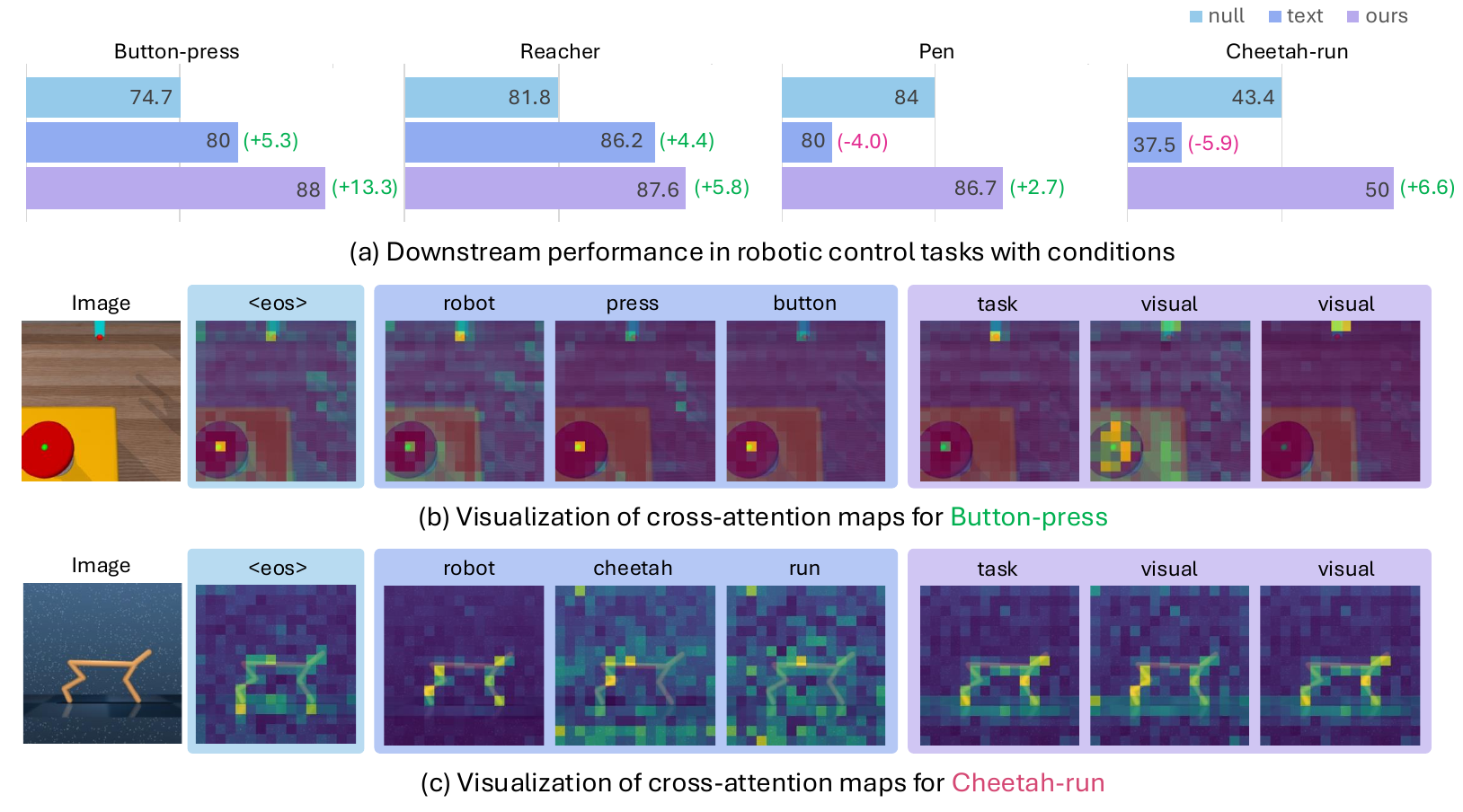}
  \vspace{-15pt}
  \caption{\textbf{Case study.} \textbf{(a)} We find that text conditions can be disadvantageous in some control tasks. \textbf{(b)} For \textit{Button-press}, the cross-attention maps (e.g., for \textit{button}, \textit{press}) are well-grounded to relevant image regions. \textbf{(c)} In contrast, for \textit{Cheetah-run}, the attention maps (e.g., for \textit{cheetah}, \textit{run}) are noisy, which presumably leads to a decline in performance. Nonetheless, our approach of using task and visual tokens (\textbf{\S~\ref{method}}) achieves consistent gains across all tasks, with its cross-attention maps capturing diverse regions of the image relevant to the downstream task.}
  \label{fig:case}
    \vspace{-1em}
\end{figure*}

In this work, we explore conditional diffusion models to generate visual representations for robotic control, aiming to overcome the limitations of task-agnostic approaches.
While pre-trained visual representations have been paramount to advancements in control, the standard approach of deploying the same frozen representation across various tasks often fails to adapt to their specific requirements, causing performance to fluctuate significantly~\citep{majumdar2023we}. We aim to address this limitation by leveraging text-to-image diffusion models, which have successfully handled diverse visual tasks in a \textbf{task-adaptive} manner using well-designed textual prompts as conditions. Our goal is therefore to explore effective ways to condition diffusion models for control, as illustrated in Fig.~\ref{fig:motivation}. 

However, we find that \textbf{text conditions are ineffective in robotic control environments} (\textbf{\S~\ref{text}}), as using captions generated from vision-language models yields insignificant gains, or even degrades performance. An in-depth inspection of the cross-attention maps reveals the underlying reason for this failure - in tasks where performance degrades, the diffusion model struggles to correctly associate words with their corresponding image regions. This underscores the need for alternatives to text descriptions and for careful consideration when devising conditions specifically for robotic control.

Consequently, we discuss what do we need for effectively conditioning diffusion models in robotic control (\textbf{\S~\ref{discussion}}). By their nature, control tasks involve video frames with fine-grained movements of agents and objects. Relying solely on textual conditions would necessitate generating a highly detailed, frame-by-frame description of the specific agent parts relevant to the current action—a challenging and often impractical task. Therefore, we posit that we should \textbf{incorporate visual information} for effective conditions to capture the fine-grained details of each frame.

\begin{figure*}[t]
  \centering
  \includegraphics[width=.8\linewidth]{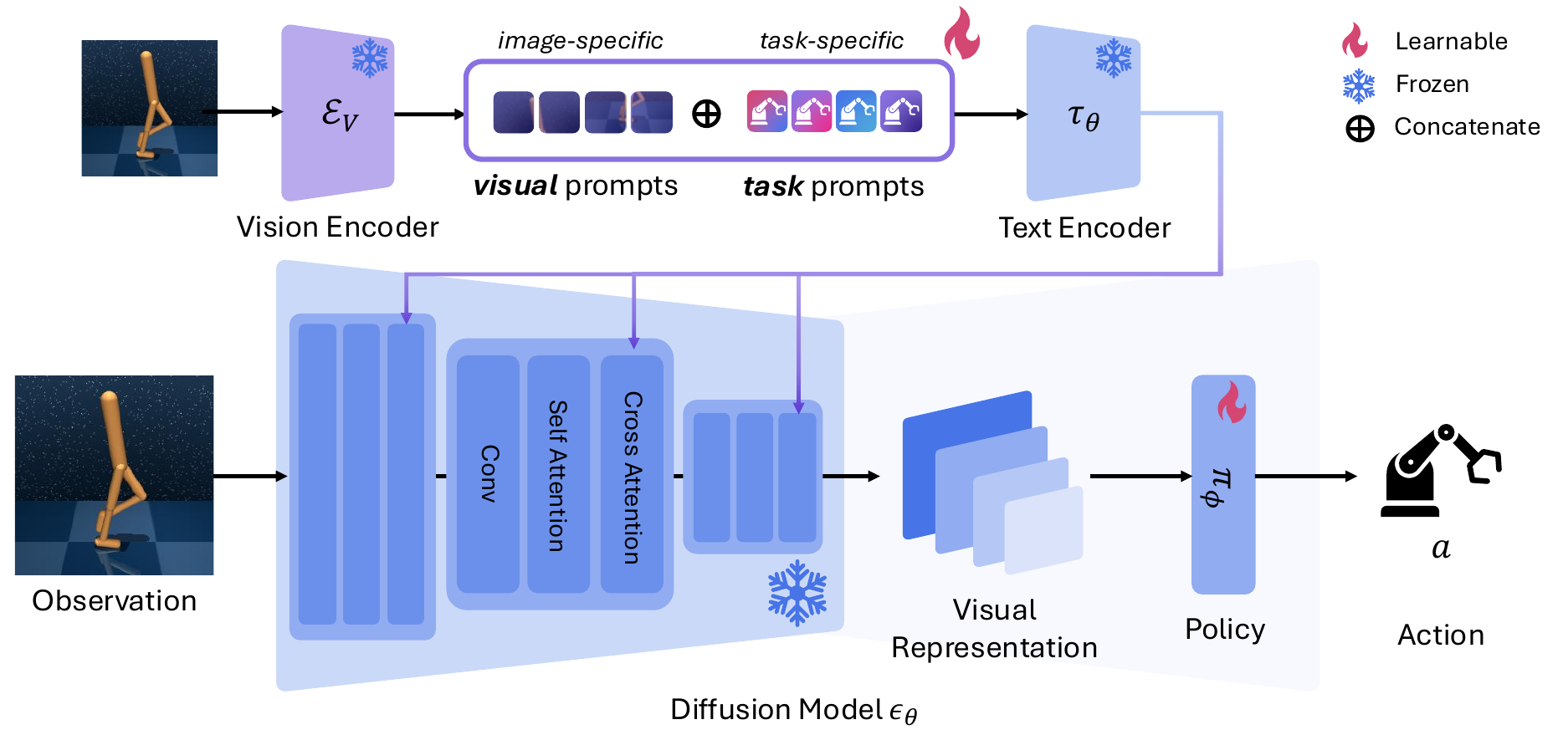}
  \caption{\textbf{Proposed framework.} We propose \textbf{\ours}, a framework for learning \textbf{task} and \textbf{visual} prompts to condition diffusion models in robotic control. Specifically, we utilize the features from the downsampling blocks and the bottleneck block of Stable Diffusion~\citep{rombach2022high} to extract visual representations conditioned on our input, which are then fed to the policy network for predicting the action.}
  \label{fig:framework}

  \vspace{-10pt}
\end{figure*}

\subsection{Exploring textual conditions for robotic control} 
\label{text}
To obtain textual descriptions of control environments, we devise a baseline by prompting a state-of-the-art vision-language model, Gemini 2.5~\citep{comanici2025gemini}, to generate descriptions of these tasks. 
The full text descriptions are provided in Section~\ref{sec:prompt} in the appendix. 
For our analysis, we compare the null ($\varnothing$) condition—implemented as an empty string with only \texttt{<eos>} and \texttt{<bos>} tokens—and the text condition in downstream control tasks. However, as observed in Fig.~\ref{fig:case}(a), the results are mixed: while text conditions benefit some tasks (\textit{e.g.}, Button-press, Reacher), they degrade performance in others (\textit{e.g.}, Cheetah-run).

To take a deeper look, in Fig.~\ref{fig:case}(b), we visualize the cross-attention maps for \textit{Button-press}, a task where text conditions show noticeable gains. For words such as \textit{press} or \textit{button}, the cross-attention maps are well-associated with the relevant regions within the image. These results are similar to what is expected from text conditions in other visual perception tasks like semantic segmentation~\citep{zhao2023unleashing}, which verifies the potential of using conditions in control tasks.

However, in Fig.~\ref{fig:case}(c), we observe the opposite for \textit{Cheetah-run}, where words like \textit{cheetah} or \textit{run} show noisy cross-attention maps. The \texttt{<eos>} token of the null condition is already roughly grounded to the salient object, the agent in this case, which explains how a sub-optimal text condition can degrade performance to be even worse than the null condition. We primarily attribute the failure of text conditions, despite being generated from a state-of-the-art vision-language model, to the domain gap between real-world images and simulated control environments. This finding highlights the need for careful consideration when devising conditions in robotic control and motivates the exploration of alternatives to text descriptions for representing the task.

\subsection{What do we need as conditions in control?}
\label{discussion}
In order to devise effective conditions, we discuss the characteristics of robotic control tasks and contrast with other vision-based tasks, such as semantic segmentation. A primary distinction is that control tasks operate on video streams rather than static images. Consequently, a logical approach would be to generate a unique condition for each frame~\citep{hong2024large}, allowing the representation to adapt to the changing visual state of the environment. For instance, instructing an agent to walk requires a sequence of distinct commands (\textit{e.g.}, move the left foot, then the right). Similarly, an effective condition should vary across frames to guide such dynamic behaviors. However, generating high-quality text descriptions on a frame-by-frame basis would not only be challenging but would also inherit the same grounding limitations discussed previously.

In this regard, we hypothesize that to account for this dynamic adaptability, conditions should incorporate \textbf{visual} information from each frame. While diffusion models like Stable Diffusion are typically trained on text, several approaches exist for incorporating visual information, either by introducing features from external vision encoders~\citep{li2024sd4match, patni2024ecodepth} or by optimizing specialized text tokens to represent visual concepts~\citep{kondapaneni2024text, kim2025diffusion}. These existing methods, such as TADP~\cite{kondapaneni2024text} however, tend to embed the global representation into the condition or require additional optimization steps to acquire specialized tokens. Since our goal is to enable the recognition of fine-grained regions within each frame, we consider that adopting global representations and extra optimization steps should be avoided to facilitate effective frame-wise conditioning.

\section{\ours: Conditioning diffusion models for robotic control}
\label{method}
Based on our observations, we present \ours, a simple yet effective approach for learning conditions for diffusion models in robotic control. We design our conditions to adapt to the control environment, preventing erroneous grounding, while simultaneously incorporating visual information to capture dynamic details. To achieve this with minimal overhead during downstream policy learning, we formulate these conditions as learnable prompts~\citep{gal2022image}. Specifically, we introduce learnable \textbf{task} and \textbf{visual} prompts that integrate task-specific implicit descriptions with frame-level visual information, as described in detail below.

\paragraph{Task prompts.}
Recalling that text conditions show potential when well-grounded to task-relevant regions, we design our task prompt to capture objects or areas that are critical to solving the downstream task. Therefore, we adopt a direct approach of learning the text as implicit words~\citep{zhou2022learning} within the downstream task to minimize erroneous grounding. To achieve this, we implement task prompts as learnable parameters that are shared across all observations during training. We find that this allows the task prompts to implicitly learn to focus on relevant regions, as shown in Fig.~\ref{fig:case}(b,c), where the cross-attention maps simultaneously highlight both the button and the robot arm in \textit{Button-press} and the agent in \textit{Cheetah-run}.

\paragraph{Visual prompts.}
Furthermore, to incorporate visual information into the conditions, we adopt a vision encoder $\mathcal{E}_V$ to leverage its visual representation as prompts. Specifically, we utilize the dense visual representations from $\mathcal{E}_V$, rather than global representations, and project them through a small convolutional layer to complement the task prompts. This focus on dense features provides the fine-grained, localized information necessary for control tasks. As visualized in Fig.~\ref{fig:case}(c), the resulting attention from the visual prompts highlights various regions in detail, such as distinguishing between the front and back legs of the agent.

\paragraph{Policy learning objective.}
We learn the prompts by directly optimizing for the behavior cloning objective in downstream policy learning, as presented in Fig.~\ref{fig:framework}.
Let $\pi_\phi(\cdot)$ be the policy network with parameters $\phi$ that takes the visual state representations derived by the diffusion model and outputs actions. Given sequences of $T_o$ observations $\{I_{o}^i\}_{o=0}^{T_o}$ and actions $\{a_{o}^i\}_{o=0}^{T_o}$ from the $i$-th trajectory, we predict each action and train both the policy network $\pi_\phi(\cdot)$, task prompts $p_t$ and visual prompts $p_t$ by the behavior cloning loss:
\begin{equation}
    \mathcal{L_{\text{BC}(\phi, \mathbf{p})}} = \sum^N_{i=1} \sum_{o}  ||\pi_\phi( \epsilon_\theta(z_{t}, t; \mathcal{C^*})) - a_{o}^i||,
\end{equation}
where $z_{t}=\sqrt{\bar{\alpha}_{t}}\mathcal{E}(I_{o}^i) + \sqrt{1-\bar{\alpha}_{t}}\epsilon$, and condition $\mathcal{C}^* = \tau_\theta(p_t;p_v)$ is derived from the text encoder $\tau_\theta$ with task prompt $p_v$ and visual prompt $p_v$ as the input. We find that $p_v$ and $p_t$ can be both learned with the behavior cloning loss in downstream policy learning.

\paragraph{Discussion.} Since our method learns prompts during downstream training, one might argue that task-adaptive representations could alternatively be achieved by fine-tuning the diffusion model itself. However, this would require a fully-tuned model to be stored for each task. ORCA, on the other hand, simultaneously learns task prompts and visual prompts from the current visual state to derive task-specific conditions during the downstream training, without the need for additional optimization as of in TADP~\cite{kondapaneni2024text}. This allows the prompting modules to be swapped for handling different tasks, as the diffusion model itself is not modified. Furthermore, we find the fine-tuning approach—exemplified by SCR~\citep{gupta2024pre}—results in significantly degraded performance, with success rates collapse by over 80\% compared to the frozen counterpart (\S~\ref{analysis}), as full fine-tuning on limited imitation learning data leads to severe overfitting~\citep{majumdar2023we, parisi2022unsurprising}.
\section{Experiments}
\begin{table*}[t]
  \caption{\textbf{Experimental results on vision-based robot policy learning on DeepMind Control~\cite{tassa2018deepmind} and MetaWorld~\cite{metaworld}}. We report the normalized score for DeepMind Control and success rates (\%) for MetaWorld, averaged over three seeds with standard deviation.}
  \vspace{-5pt}
  \label{tab:combined_results}
  \centering
  \resizebox{\linewidth}{!}{
  \begin{tabular}{l|c|ccccc|c|ccccc|c}
    \toprule
    \multirow{2}{*}{Methods} & \multirow{2}{*}{Backbone} & \multicolumn{5}{c|}{\text{DeepMind Control}} & &\multicolumn{5}{c|}{\text{MetaWorld}} &\\
     & & Stand & Walk & Reacher & Cheetah & Finger & Mean & Assembly & Bin-picking & Button & Drawer & Hammer & Mean \\
    \midrule
    \midrule
    CLIP~\cite{radford2021learning} & ViT-L/16 
    & 87.3 \footnotesize{$\pm$ 2.4} & 58.3 \footnotesize{$\pm$ 4.4} & 54.5 \footnotesize{$\pm$ 4.6} & 29.9 \footnotesize{$\pm$ 5.6} & 67.5 \footnotesize{$\pm$ 2.1} & 59.5 
    & \phantom{0}85.3 \footnotesize{$\pm$ 12.2} & 69.3 \footnotesize{$\pm$ 8.3} & \phantom{0}60.0 \footnotesize{$\pm$ 13.9} & \textbf{100.0} \footnotesize{$\pm$ 0.0}\phantom{0} & 92.0 \footnotesize{$\pm$ 8.0} & 81.3 \\
    
    VC-1~\cite{majumdar2023we} & ViT-L/16 
    & 86.1 \footnotesize{$\pm$ 0.9} & 54.3 \footnotesize{$\pm$ 6.6} & 18.3 \footnotesize{$\pm$ 2.4} & 40.9 \footnotesize{$\pm$ 2.7} & 65.7 \footnotesize{$\pm$ 1.1} & 53.1 
    & \text{93.3} \footnotesize{$\pm$ 6.1} & \phantom{0}61.3 \footnotesize{$\pm$ 12.2} & 73.3 \footnotesize{$\pm$ 8.3} & \textbf{100.0} \footnotesize{$\pm$ 0.0}\phantom{0} & 93.3 \footnotesize{$\pm$ 6.1} & 84.2 \\
    
    SCR~\cite{gupta2024pre} & SD 1.5 
    & 85.5 \footnotesize{$\pm$ 2.6} & 64.3 \footnotesize{$\pm$ 3.5} & 81.8 \footnotesize{$\pm$ 9.9} & 43.4 \footnotesize{$\pm$ 6.4} & 66.6 \footnotesize{$\pm$ 2.7} & 68.3 
    & 92.0 \footnotesize{$\pm$ 6.9} & \text{86.7} \footnotesize{$\pm$ 4.6} & \phantom{0}\text{74.7} \footnotesize{$\pm$ 12.9} & \textbf{100.0} \footnotesize{$\pm$ 0.0}\phantom{0} & \textbf{98.7} \footnotesize{$\pm$ 2.3} & 90.4 \\

    Text (Simple) & SD 1.5 
    & 87.6 \footnotesize{$\pm$ 4.6} & 67.9 \footnotesize{$\pm$ 4.6} & 84.3 \footnotesize{$\pm$ 4.6} & 38.8 \footnotesize{$\pm$ 5.9} & 66.7 \footnotesize{$\pm$ 0.2} & 69.1 
    & \text{97.3} \footnotesize{$\pm$ 2.3} & \text{85.3} \footnotesize{$\pm$ 2,3} & \text{78.7} \footnotesize{$\pm$ 2,3} & \textbf{100.0} \footnotesize{$\pm$ 0.0}\phantom{0} & \text{96.0} \footnotesize{$\pm$ 6.9} & \text{91.5} \\

    Text (Caption) & SD 1.5 
    & 87.2 \footnotesize{$\pm$ 4.5} & 68.3 \footnotesize{$\pm$ 5.9} & 86.2 \footnotesize{$\pm$ 1.9} & 37.5 \footnotesize{$\pm$ 2.6} & 65.1 \footnotesize{$\pm$ 1.8} & 68.9 
    & \text{96.0} \footnotesize{$\pm$ 4.0} & \text{88.0} \footnotesize{$\pm$ 6.9} & \text{80.0} \footnotesize{$\pm$ 8.0} & \textbf{100.0} \footnotesize{$\pm$ 0.0}\phantom{0} & \textbf{98.7} \footnotesize{$\pm$ 2.3} & \text{92.5} \\

    CoOp~\cite{zhou2022learning} & SD 1.5 
    & 87.2 \footnotesize{$\pm$ 2.2} & 67.8 \footnotesize{$\pm$ 6.4} & 87.1 \footnotesize{$\pm$ 5.9} & 45.0 \footnotesize{$\pm$ 6.4} & 65.9 \footnotesize{$\pm$ 1.0} & 70.6 
    & \text{96.0} \footnotesize{$\pm$ 4.0} & \text{89.3} \footnotesize{$\pm$ 2,3} & \text{81.3} \footnotesize{$\pm$ 6.1} & \textbf{100.0} \footnotesize{$\pm$ 0.0}\phantom{0} & \text{96.0} \footnotesize{$\pm$ 6.9} & 92.5 \\

    TADP~\cite{kondapaneni2024text} & SD 1.5 
    & 89.0 \footnotesize{$\pm$ 2.9} & 69.9 \footnotesize{$\pm$ 7.9} & 86.6 \footnotesize{$\pm$ 5.6} & 41.1 \footnotesize{$\pm$ 3.9} & 66.9 \footnotesize{$\pm$ 0.2} & 70.7 
    & \text{96.0} \footnotesize{$\pm$ 4.0} & \textbf{90.7} \footnotesize{$\pm$ 4.6} & \phantom{0}\text{80.0} \footnotesize{$\pm$ 10.6} & \textbf{100.0} \footnotesize{$\pm$ 0.0}\phantom{0} & \text{96.0} \footnotesize{$\pm$ 4.0} & 93.1 \\
\midrule
    \textbf{\ours(Ours)} & SD 1.5 
    & \textbf{89.1} \footnotesize{$\pm$ 1.8} & \textbf{76.9} \footnotesize{$\pm$ 4.0} & \textbf{87.6} \footnotesize{$\pm$ 2.9} & \textbf{50.0} \footnotesize{$\pm$ 8.4} & \textbf{68.0} \footnotesize{$\pm$ 1.0} & \textbf{74.3} 
    & \textbf{98.7} \footnotesize{$\pm$ 2.3} & \textbf{90.7} \footnotesize{$\pm$ 4.6} & \textbf{88.0} \footnotesize{$\pm$ 6.9} & \textbf{100.0} \footnotesize{$\pm$ 0.0}\phantom{0} & \textbf{98.7} \footnotesize{$\pm$ 2.3} & \textbf{95.2} \\
    \bottomrule
  \end{tabular}}
\vspace{-10pt}
\end{table*}

\begin{figure}
  \centering
  \vspace{-10pt}
  \includegraphics[width=1.0\linewidth]{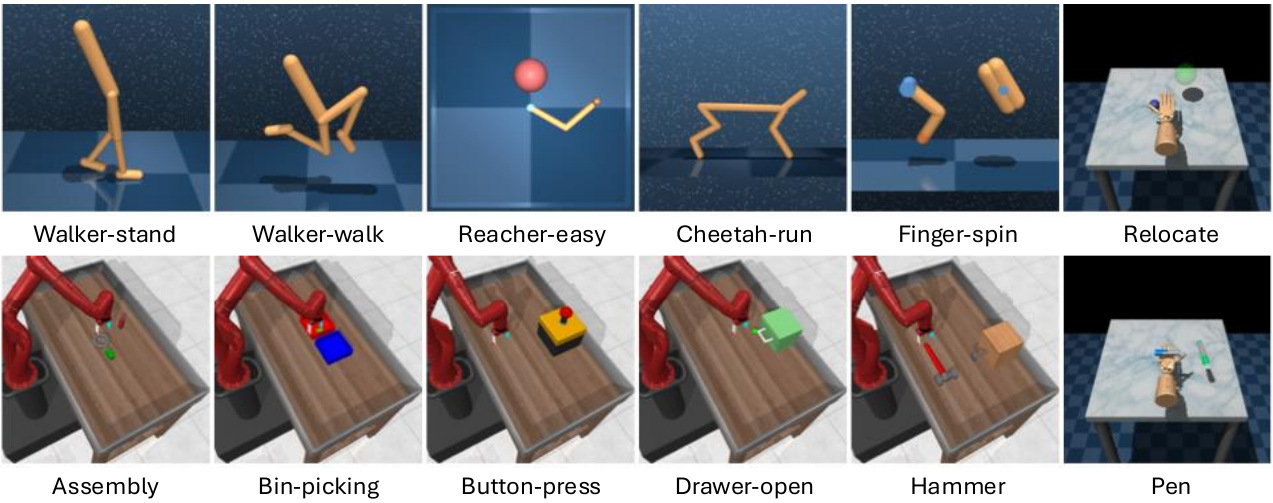}
  \caption{\textbf{Visualization of evaluation tasks.} We utilize MuJoCo~\cite{todorov2012mujoco} tasks for evaluation, with 5 tasks from MetaWorld~\cite{metaworld}, 5 tasks from DeepMind Control~\cite{tassa2018deepmind}, and 2 tasks from Adroit~\cite{adroit}.}
  \label{fig:task}

  \vspace{-15pt}
\end{figure}

\subsection{Evaluation suites} \label{sec:eval}
We conduct experiments on three widely-used vision-based robot learning benchmarks with the total of 12 tasks following VC-1~\citep{majumdar2023we}, as shown in Fig.~\ref{fig:task}.

\textbf{DeepMind Control (DMC)}~\citep{tassa2018deepmind} is a set of continuous control tasks with simulated robots. 
We use five imitation learning cases: \textit{Walker-stand}, \textit{Walker-walk}, \textit{Reacher-easy}, \textit{Cheetah-run}, and \textit{Finger-spin}. We report the normalized scores for all tasks.

\textbf{MetaWorld}~\citep{metaworld} is a suite of simulated robotic manipulation tasks with a Sawyer robot arm. We focus on a subset of five representative tasks: \textit{Assembly}, \textit{Bin-picking}, \textit{Button-press-topdown}, \textit{Drawer-open}, and \textit{Hammer}. We measure the best success rates among the online evaluation trials. 

\textbf{Adroit}~\citep{adroit} is an imitation learning benchmark in a simulated environment, consisting of dexterous manipulation tasks that require an agent to control a 28-DoF anthropomorphic hand. We focus on \textit{Relocate} and \textit{Pen}, and measure the best success rates among the online evaluation trials.

\subsection{Implementation details} \label{sec:details}
\noindent\textbf{Diffusion model and conditions.} 
We employ Stable Diffusion v1.5~\citep{rombach2022high} as the diffusion model. For extracting visual representation from observations, we leverage the features from the downsampling blocks and the bottleneck block in the diffusion U-Net and forward through a compression layer~\citep{yadav2023ovrl}. We set the timestep $t=0$, the length of task tokens $l_t=4$, and the length of visual tokens $l_v=16$, where all learnable parameters are randomly initialized. For $\mathcal{E}_V$, we employ pre-trained DINOv2~\citep{oquab2023dinov2}. Further implementation details are presented in the appendix.

\noindent\textbf{Vision-based robot policy learning.} We consider two, five, and five demonstrations from Adroit, DeepMind Control (DMC), and MetaWorld, respectively, where proprioceptive data is utilized except for the DMC benchmark. We mainly follow the experimental setups in VC-1~\citep{majumdar2023we} except that we employ a compression layer for all baselines for fair comparison. For each task, we train the agent for 100 epochs, with a periodic online evaluation in the simulated environment every 10 epochs.

\subsection{Baselines}
For baselines, we consider CLIP~\cite{radford2021learning} and VC-1~\cite{majumdar2023we} as widely adopted \textit{task-agnostic} baselines. Furthermore, we construct 5 baselines based on diffusion models to explore conditions in robotic control, with details in the appendix.
\begin{itemize}
    \item \textbf{SCR~\cite{gupta2024pre}} is a \textit{task-agnostic} baseline, which is one of the first works to introduce Stable Diffusion into control tasks. SCR can be considered as an unconditional baseline, where it employs the null($\varnothing$) condition for all tasks.
    \item \textbf{Text (Simple/Caption)} are \textit{task-adaptive} baselines using text conditions, where the simple variant uses the task names defined in the evaluation suite, and caption variant leverages descriptions generated from Gemini 2.5~\citep{comanici2025gemini}.
    \item \textbf{CoOp}~\citep{zhou2022learning} extends on Text (Simple) by implementing \textit{learnable} prefix tokens to the text.  CoOp can be considered as a less flexible variant to the task prompt as the task name is fixed in the prompt.
    \item \textbf{TADP}~\citep{kondapaneni2024text} extends on Text (Caption) by appending a special token $S^*$ that encapsulates the \textit{visual} style information, separately optimized through Textual Inversion~\citep{gal2022image}. Hence, we can consider TADP as a baseline with visual information from only a single, fixed image.
\end{itemize}

\begin{table}[t]
  \caption{\textbf{Experimental results on vision-based robot policy learning on Adroit~\cite{adroit}}. We report the success rates ($\%$) averaged over three seeds with their standard deviation.}
  \vspace{-5pt}
  \label{tab:adroit}
  \centering
  \footnotesize
  \begin{tabular}{l|c|cc|c}
    \toprule
\multirow{2}{*}{Methods} & \multirow{2}{*}{Backbone}  & \multicolumn{2}{c|}{Adroit} & \\
 &  & Pen & Relocate  & Mean\\
    \midrule
    \midrule
    
CLIP~\cite{radford2021learning} &   ViT-L/16 &  58.7 \footnotesize{$\pm$ 2.3} &  \textbf{44.0} \footnotesize{$\pm$ 4.0} & 51.4 \\
VC-1~\cite{majumdar2023we} & ViT-L/16 & \phantom{0}65.3 \footnotesize{$\pm$ 16.7} & 29.3 \footnotesize{$\pm$ 8.3} & 47.3 \\
SCR~\cite{gupta2024pre} & SD 1.5 & \text{84.0} \footnotesize{$\pm$ 4.0} & 32.0 \footnotesize{$\pm$ 4.0} & 58.0  \\
Text (Simple) & SD 1.5  & 80.0 \footnotesize{$\pm$ 6.9} & 34.7 \footnotesize{$\pm$ 6.1} & 57.3 \\
Text (Caption) & SD 1.5  & 80.0 \footnotesize{$\pm$ 4.0} & 34.7 \footnotesize{$\pm$ 4.6} & 57.3  \\
CoOp~\cite{zhou2022learning} & SD 1.5 & 82.7 \footnotesize{$\pm$ 6.1} & 33.3 \footnotesize{$\pm$ 6.1} & 58.0  \\
TADP~\cite{kondapaneni2024text} & SD 1.5  & \text{81.3} \footnotesize{$\pm$ 6.1}& 33.3 \footnotesize{$\pm$ 8.3}& 57.3\\
\midrule
\textbf{\ours(Ours)}& SD 1.5 & \textbf{86.7} \footnotesize{$\pm$ 2.3} & \textbf{44.0} \footnotesize{$\pm$ 4.0} & \textbf{65.3}  \\
  \bottomrule
  \end{tabular}
\vspace{-10pt}
\end{table}
\subsection{Main results} \label{results}
\noindent\textbf{Quantitative results.} We report experimental results from DMC and MetaWorld in Table~\ref{tab:combined_results}, and Adroit in Table~\ref{tab:adroit}. Among the task-agnostic baselines, while SCR performs best overall, we observe that VC-1 and CLIP outperform it in certain tasks. This highlights a fundamental limitation of such approaches: due to their task-agnostic nature, no single representation is guaranteed to excel across all tasks. In contrast, across all 12 tasks in the 3 evaluation suites, \ours establishes the new state-of-the-art, outperforming all baselines by a significant margin. 

Furthermore, we observe that more advanced task-adaptive baselines, CoOp and TADP, generally outperform text conditions. This confirms our hypothesis that incorporating visual information is beneficial, given from the results of TADP which utilizes visual information in a limited manner with a global observation of the task. Nonetheless, since both methods were not designed for robotic control tasks, their effectiveness is limited as shown by their minimal gains on DMC and Adroit. In contrast, our method show solid improvements across all tasks.

\subsection{Analysis} \label{analysis}
\begin{table}[t]
  \caption{\textbf{Comparison to fine-tuning}. For comparison, we fine-tune VC-1~\cite{majumdar2023we} and SCR~\cite{gupta2024pre} on Adroit~\citep{adroit} under full-finetuning and parameter-efficient fine-tuning scenarios with RoboAdapter~\cite{sharma2023lossless} and LoRA~\cite{hu2022lora}. We report the number of learnable parameters and the performance in success rates ($\%$) averaged over three seeds with their standard deviation.}
  \vspace{-5pt}
  \label{tab:finetune}
  \centering
  \footnotesize
  \begin{tabular}{l|c|cc|c}
    \toprule
\multirow{2}{*}{Methods} & \# learn.  & \multicolumn{2}{c|}{Adroit} & \\
 & params. & Pen & Relocate  & Mean\\
    \midrule
    \midrule

VC-1~\cite{majumdar2023we} & - & \phantom{0}65.3 \footnotesize{$\pm$ 16.7} & 29.3 \footnotesize{$\pm$ 8.3} & 47.3 \\
+ Fine-tuning & 302.3M & \phantom{0}58.7 \footnotesize{$\pm$ 13.2} & \phantom{0}4.0 \footnotesize{$\pm$ 3.3} & 31.3 \\
+ RoboAdapter & \phantom{0}18.0M & 77.3 \footnotesize{$\pm$ 4.6} & 41.3 \footnotesize{$\pm$ 8.3} & 59.3 \\

\midrule

SCR~\cite{gupta2024pre} & - & \text{84.0} \footnotesize{$\pm$ 4.0} & 32.0 \footnotesize{$\pm$ 4.0} & 58.0  \\
+ Fine-tuning & 346.7M & \text{17.3} \footnotesize{$\pm$ 3.8} & \phantom{0}1.3 \footnotesize{$\pm$ 1.9} & \phantom{0}9.3  \\
+ \text{LoRA} & \phantom{00}4.6M & 77.3 \footnotesize{$\pm $6.1} & \phantom{0}42.7 \footnotesize{$\pm$ 15.1} & 60.0 \\ 

\midrule
\textbf{\ours(Ours)}& 10.6M & \textbf{86.7} \footnotesize{$\pm$ 2.3} & \textbf{44.0} \footnotesize{$\pm$ 4.0} & \textbf{65.3}  \\
  \bottomrule
  \end{tabular}
\vspace{-10pt}
\end{table}

\begin{table}[t]
  \caption{\textbf{Components analysis}. To ablate the design choices, we conduct component analysis on task prompt $p_t$ and visual prompt $p_v$ on DeepMind Control~\citep{tassa2018deepmind}. We report the normalized score averaged over three seeds with its standard deviation.}
  \vspace{-5pt}
  \label{tab:ablations}
  \centering
  \resizebox{\linewidth}{!}{
  \begin{tabular}{cc|ccccc|c}
    \toprule
\multicolumn{2}{c|}{} & \multicolumn{5}{c|}{DeepMind Control} & \\
$p_t$ & $p_v$ &  Stand & Walk & Reacher & Cheetah & Finger & Mean\\
    \midrule
    \midrule

& & 85.5 \footnotesize{$\pm$ 2.6} & 64.3 \footnotesize{$\pm$ 3.5}  & 81.8 \footnotesize{$\pm$ 1.7} & 43.4 \footnotesize{$\pm$ 4.4}\phantom{0}  & 66.6 \footnotesize{$\pm$ 2.7} & 68.3 \\

\checkmark& & 83.6 \footnotesize{$\pm$ 3.2} & \text{71.4} \footnotesize{$\pm$ 3.5}  & 86.7 \footnotesize{$\pm$ 6.6} & 38.9 \footnotesize{$\pm$ 10.1}  & \textbf{68.2} \footnotesize{$\pm$ 1.2} & 69.8 \\

&\checkmark &\text{85.9} \footnotesize{$\pm$ 2.7} & 71.1 \footnotesize{$\pm$ 2.3}  & 87.3 \footnotesize{$\pm$ 5.5} & 42.0 \footnotesize{$\pm$ 10.4}  & 66.1 \footnotesize{$\pm$ 1.0} & \text{70.5} \\

\checkmark&\checkmark & \textbf{89.1} \footnotesize{$\pm$ 2.3} & \textbf{76.9} \footnotesize{$\pm$ 4.0}  & \textbf{87.6} \footnotesize{$\pm$ 2.9}   & \textbf{50.0} \footnotesize{$\pm$ 8.4}\phantom{0}  & \text{68.0} \footnotesize{$\pm$ 1.0} & \textbf{74.3} \\

\bottomrule

  \end{tabular}}
\vspace{-10pt}
\end{table}

\noindent\textbf{Comparison with downstream fine-tuning.}
To compare learning prompts to downstream fine-tuning, we fine-tune VC-1~\citep{majumdar2023we} and SCR~\citep{gupta2024pre} on Adroit~\citep{adroit}, evaluating both full fine-tuning and parameter-efficient fine-tuning with adapter modules. For the parameter-efficient methods, we use RoboAdapter~\cite{sharma2023lossless} for ViT-based VC-1 and LoRA~\cite{hu2022lora} for diffusion-based SCR. As shown in Table~\ref{tab:finetune}, full fine-tuning requires approximately $30\times$ more parameters but proves ineffective, completely deteriorating SCR's performance. Although parameter-efficient methods mitigate this, ORCA significantly boosts performance while maintaining a comparable parameter count, highlighting the superior efficiency and efficacy of our condition-based approach.

\noindent\textbf{Ablation study on each component.}
In Table ~\ref{tab:ablations}, we conduct component analysis by ablating task prompt $p_t$ and visual prompt $p_v$ respectively. Notably, we observe that when employed individually, task and visual prompts can show divergent behavior across different tasks. This could suggest that each tasks focus in different aspects of the scene, such as \textit{Reacher-easy} focusing more in visual details as it benefits more from visual prompts compared to text prompts. Nonetheless, when fully incorporating both $p_v$ and $p_t$, we show consistent gains across all tasks.

\noindent\textbf{Ablation study on layer selection.}
In Table ~\ref{tab:layer}, we evaluate the diffusion model in different layers. We can observe that the early layers tend to perform better than the latter upsampling layers. Therefore, we concatenate the best-performing layers (\texttt{down\_1-3,mid}), which yields the best overall performance.

\begin{table}[t]
\small
\caption{\textbf{Ablation study on layer selection}. We evaluate individual layers of the diffusion U-Net by reporting layer-wise performance on DeepMind Control~\citep{tassa2018deepmind}. \texttt{d} + \texttt{m} refers to concatenating \texttt{down} and \texttt{mid} blocks. We report the normalized score averaged over three seeds with its standard deviation.} \label{tab:layer}
\vspace{-5pt}
  
  \centering
  \resizebox{\linewidth}{!}{
  \begin{tabular}{l|ccccc|c}
    \toprule
 & \multicolumn{5}{c|}{DeepMind Control} &\\
Layer &  Stand & Walk & Reacher & Cheetah & Finger & Mean\\
    \midrule
    \midrule
\texttt{down\_1} & 86.3 \footnotesize{$\pm$ 2.1} & 65.5 \footnotesize{$\pm$ 1.1} & \text{82.1} \footnotesize{$\pm$ 3.7} & \text{40.8} \footnotesize{$\pm$ 1.1} & \text{67.6} \footnotesize{$\pm$ 0.3} & \text{68.4} \\

\texttt{down\_2} & \textbf{89.3} \footnotesize{$\pm$ 1.2} & 68.3 \footnotesize{$\pm$ 2.7} & \phantom{0}70.0 \footnotesize{$\pm$ 18.8} & 31.2 \footnotesize{$\pm$ 2.6} & 67.0 \footnotesize{$\pm$ 1.0} & 65.1 \\

\texttt{down\_3} & 86.2 \footnotesize{$\pm$ 4.3} & \text{73.3} \footnotesize{$\pm$ 3.9} & \text{75.3} \footnotesize{$\pm$ 8.1} & 36.0 \footnotesize{$\pm$ 4.8} & 67.0 \footnotesize{$\pm$ 0.5} & 67.5 \\

\texttt{mid} & 88.3 \footnotesize{$\pm$ 4.9} & 70.4 \footnotesize{$\pm$ 1.3} & 62.3 \footnotesize{$\pm$ 1.1} & 35.0 \footnotesize{$\pm$ 4.7} & 67.2 \footnotesize{$\pm$ 0.6} & 64.6\\

\texttt{up\_0} & 82.8 \footnotesize{$\pm$ 2.6} & 71.7 \footnotesize{$\pm$ 5.9} & 45.3 \footnotesize{$\pm$ 4.0} & 28.5 \footnotesize{$\pm$ 1.8} & 67.2 \footnotesize{$\pm$ 0.6} & 59.0\\

\texttt{up\_1} & 79.5 \footnotesize{$\pm$ 4.5} &  \phantom{0}60.3 \footnotesize{$\pm$ 16.1} & 55.9 \footnotesize{$\pm$ 5.2} & 39.9 \footnotesize{$\pm$ 7.0} & 66.4 \footnotesize{$\pm$ 0.4} & 60.4\\

\texttt{up\_2} & 70.4 \footnotesize{$\pm$ 4.5} & 39.1 \footnotesize{$\pm$ 3.3} & 41.0 \footnotesize{$\pm$ 7.0} & 30.9 \footnotesize{$\pm$ 3.1} & 67.7 \footnotesize{$\pm$ 1.0} & 49.7 \\

\midrule

\texttt{d} + \texttt{m} & \text{89.1} \footnotesize{$\pm$ 1.8} & \textbf{76.9} \footnotesize{$\pm$ 4.0}  & \textbf{87.6} \footnotesize{$\pm$ 2.9}   & \textbf{50.0} \footnotesize{$\pm$ 8.4}  & \textbf{68.0} \footnotesize{$\pm$ 1.0} & \textbf{74.3} \\

\bottomrule
  \end{tabular}}
  \vspace{-10pt}
\end{table}

\begin{figure}[t]
  \centering
  \includegraphics[width=1.0\linewidth]{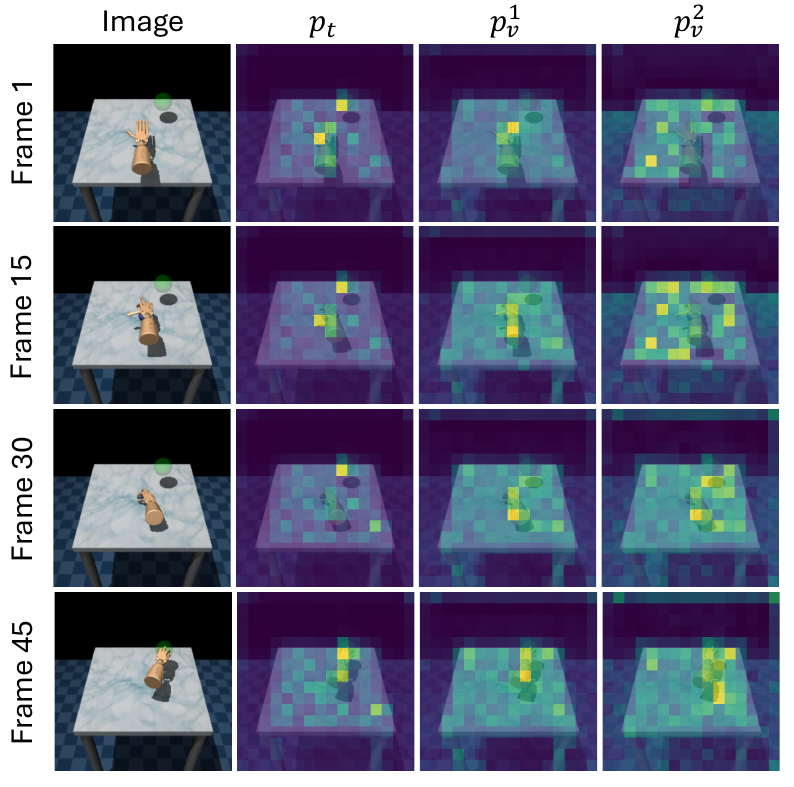}
  \vspace{-20pt}
  \caption{\textbf{Cross-attention visualization for task/visual prompts.} We visualize the cross-attention maps for task prompt $p_t$ and visual prompts $p_v^1$ and $p_v^2$ in across frames in the \textit{Relocate} task. }
  \label{fig:visualization}

  \vspace{-15pt}
\end{figure}

\noindent\textbf{Visualization of task and visual prompts.}
In Fig.~\ref{fig:visualization}, we visualize the cross-attention maps for our task prompt $p_t$ and visual prompts, $p_v^1$ and $p_v^2$, on \textit{Relocate}. In this task, a robot hand first picks up a blue ball from a table (Frames 1-30) and then moves it to the location of a green sphere (Frames 30-45). As discussed in \textbf{\S~\ref{method}}, we observe that the task prompt consistently captures regions relevant to the overall goal, namely the robot hand and the target green sphere. Conversely, the visual prompts exhibit more dynamic behaviors. While $p_v^1$ tends to focus on the hand, $p_v^2$ interestingly attends to the table as the hand moves downward to pick up the ball, then shifts its focus to the hand as it lifts off and moves toward the target, suggesting that it has learned to capture task-relevant movements.

\noindent\textbf{Visualization of task and visual prompts.}
In Fig.~\ref{fig:visualization}, we visualize the cross-attention maps for our task prompt $p_t$ and two different tokens from the visual prompt, $p_v^1$ and $p_v^2$, on \textit{Relocate}. In this task, a robot hand first picks up a blue ball from a table (Frames 1-30) and then moves it to the location of a green sphere (Frames 30-45). As discussed in \textbf{\S~\ref{method}}, we observe that the task prompt consistently captures regions relevant to the overall goal, namely the robot hand and the target green sphere. Conversely, the visual prompts exhibit more dynamic behaviors. While $p_v^1$ tends to focus on the hand, $p_v^2$ interestingly attends to the table as the hand moves downward to pick up the ball, then shifts its focus to the hand as it lifts off and moves toward the target, suggesting that it has learned to capture task-relevant movements.

\section{Conclusion}

In this work, we introduced \ours, a framework for bridging text-to-image diffusion models for robotic control to generate task-adaptive visual representations. We identified the limitations of conventional text prompts, and we proposed a simple yet effective method using learnable task and visual prompts. \ours achieves state-of-the-art performance, highlighting the importance of task-adaptive representations and the vast potential of properly conditioned diffusion models for robotic control.

\setcounter{page}{1}

\setcounter{table}{0}
\renewcommand{\thetable}{\Alph{table}}
\setcounter{figure}{0}
\renewcommand{\thefigure}{\Alph{figure}}
\setcounter{section}{0}
\renewcommand\thesection{\Alph{section}}

\maketitlesupplementary

\section*{Contents}
\begin{itemize}
    \item \textbf{\S\ref{sec:supp_analysis}: Further analysis}
    \begin{itemize}
        \item \S\ref{sec:svd}: Results with different diffusion backbones
        \item \S\ref{sec:libero}: Results on LIBERO-Long benchmark
        \item \S\ref{sec:visual_prompt}: Ablation on the vision encoder for visual prompt
        \item \S\ref{sec:timesteps}: Ablation on timesteps
        \item \S\ref{sec:encoders}: Comparison with stronger pre-trained encoders
        \item \S\ref{sec:efficiency}: Efficiency comparison
        \item \S\ref{sec:null}: Analysis on the null condition
        \item \S\ref{sec:supp_dis}: Further discussion
    \end{itemize}
    \item \textbf{\S\ref{sec:supp_impl}: Further implementation details}
    \begin{itemize}
        \item \S\ref{sec:prompt}: Full description of the text conditions
    \item \S\ref{sec:baselines}: Details of the baselines
        \item \S\ref{sec:supp_compression}: Implementation details of the compression layer
    \end{itemize}
    \item \textbf{\S\ref{sec:limitations}: Limitations}
    \item \textbf{\S\ref{sec:qual}: Qualitative results on robotic control tasks}
\end{itemize}

\section{Further Analysis}
\label{sec:supp_analysis}

\subsection{Results with different diffusion backbones} \label{sec:svd}

\begin{table}[h]
\vspace{-10pt}
  \caption{\textbf{Experimental results with different diffusion backbones}. The performance of imitation learning agents on Adroit~\citep{adroit} is reported. We report the success rates ($\%$) averaged over three seeds with their standard deviation.} \label{tab:backbones}
  \vspace{-5pt}
  \label{tab:adroit}
  \centering
  \footnotesize
  \resizebox{\linewidth}{!}{
  \begin{tabular}{l|c|cc|c}
    \toprule
\multirow{2}{*}{Methods} &\multirow{2}{*}{Backbone}& \multicolumn{2}{c|}{Adroit} & \\
 & & Pen & Relocate  & Mean\\
    \midrule
    \midrule
    
CLIP~\cite{radford2021learning} &   ViT-L/16~\cite{dosovitskiy2020image} &  58.7 \footnotesize{$\pm$ 2.3} &  \textbf{44.0} \footnotesize{$\pm$ 4.0} & 51.4 \\
VC-1~\cite{majumdar2023we} & ViT-L/16~\cite{dosovitskiy2020image} & \phantom{0}65.3 \footnotesize{$\pm$ 16.7} & 29.3 \footnotesize{$\pm$ 8.3} & 47.3 \\
\midrule
SCR~\cite{gupta2024pre} & SD 1.5~\cite{rombach2022high} & \text{84.0} \footnotesize{$\pm$ 4.0} & 32.0 \footnotesize{$\pm$ 4.0} & 58.0  \\
Video Diffusion & SVD~\cite{blattmann2023stable} & \text{49.3} \footnotesize{$\pm$ 3.8} & \phantom{0}2.7 \footnotesize{$\pm$ 1.9} & 26.0  \\
Diffusion Transformer & SD 3.0~\cite{esser2024scaling} & \phantom{0}\text{54.7} \footnotesize{$\pm$ 16.4} & \phantom{0}5.3 \footnotesize{$\pm$ 1.9} & 30.0  \\

\midrule
\textbf{\ours(Ours)}& SD 1.5~\cite{rombach2022high} & \textbf{86.7} \footnotesize{$\pm$ 2.3} & \textbf{44.0} \footnotesize{$\pm$ 4.0} & \textbf{65.3}  \\
  \bottomrule
  \end{tabular}}
\end{table}

To study different diffusion backbones, in Table ~\ref{tab:backbones}, we present results from a video diffusion model, Stable Video Diffusion (SVD)~\cite{blattmann2023stable} and a diffusion Transformer model, Stable Diffusion 3.0 (SD 3.0)~\cite{esser2024scaling}, which has a multi-modal diffusion Transformer (MM-DiT) as its backbone. For SVD, we use the same layer selection strategy from ours and SCR as SVD is also based on the same U-Net architecture, and use the last layer for MM-DiT. 

Interestingly, despite SVD being a dedicated video generation model, it significantly falls behind SD 1.5. While there could be several reasons, we consider this to originate from input discrepancy, where we restrict the input to 3 frames for fair comparison. This deviates from the native 8-frame setting of SVD, which likely disrupts the pre-trained knowledge. Moreover, the publicly released version of SVD does not support text conditioning, which precludes the use of text or learnable prompts without additional fine-tuning for customizing the model to accept text~\cite{hu2024video}.

On the other hand, SD 3.0, based on the MM-DiT architecture~\cite{esser2024scaling}, supports text conditioning through its joint multi-modal attention mechanism. However, we find that the DiT-based model underperforms compared to U-Net-based models, regardless of whether text conditions are provided. Given that empirical studies~\cite{kim2025seg4diff} on DiT-based models remain limited compared to the extensive research on U-Net architectures~\cite{gan2025unleashing,xu2024matters}, we believe there is significant room for improvement with DiT-based models. Nonetheless, we leave the deeper exploration of video diffusion and DiT-based models as a direction for future work.

\subsection{Results on LIBERO-Long benchmark.} \label{sec:libero}

\begin{table}[h]
    \caption{\textbf{Experimental results on tasks from LIBERO-Long}. The performance of imitation learning agents on LIBERO-Long~\cite{liu2023libero} benchmark. }
  \centering
  \resizebox{\linewidth}{!}{
  \begin{tabular}{l|ccc|c|c}
    \toprule
 & \multicolumn{4}{c|}{Single-task} & Multi-task\\
Methods & KITCHEN-3 &LIVING-1 &STUDY-1 & Mean & Mean\\
    \midrule
    \midrule

VC-1 & \phantom{0}\text{51.7} \footnotesize{$\pm$10.4} & \phantom{0}\text{43.3} \footnotesize{$\pm$10.4} & 
\phantom{0}\text{35.0} \footnotesize{$\pm$26.0} & 
43.3 & \text{26.7} \\

SigLIP & 51.7 \footnotesize{$\pm$7.6} &
20.0 \footnotesize{$\pm$5.0} &
\phantom{0}\text{38.3} \footnotesize{$\pm$23.1} &
36.7 & 16.7\\

SCR & \phantom{0}86.7 \footnotesize{$\pm$11.5} &
45.0 \footnotesize{$\pm$5.0} &
\text{40.0} \footnotesize{$\pm$8.7} &
57.2 & 23.3 \\

\midrule

\textbf{\textbf{ORCA} (Ours)} & \phantom{0}\textbf{93.3} \footnotesize{$\pm$11.5} &
\textbf{48.3} \footnotesize{$\pm$7.6}  &
\textbf{55.0} \footnotesize{$\pm$2.9} &
\textbf{65.5} & \textbf{46.6}\\
\bottomrule
  \end{tabular}}
    \label{tab:libero}
\end{table}

To investigate the efficacy of ORCA in long-horizon and language-conditioned tasks, we employed 3 tasks from LIBERO-long~\cite{liu2023libero} for evaluation. As shown in Table~\ref{tab:libero}, we can observe that ORCA surpasses baselines by a considerable margin, which confirms its applicability to various scenarios including long-horizon tasks. Furthermore, 
we report the multi-task results on the LIBERO-long benchmark, where we jointly train all 3 tasks with a shared policy. Notably, ORCA consistently outperforms the baselines by a significant margin, indicating that the visual and task prompts can be learned jointly from multiple tasks.

\subsection{Ablation on the choice of the visual encoder for visual prompt.} \label{sec:visual_prompt}

\begin{table}[h]
\caption{\textbf{Ablation on the vision encoder $\mathcal{E}_V$ for the visual prompts}. To ablate the choice of the vision encoder $\mathcal{E}_V$ for obtaining visual prompts, we provide provide results with SigLIP~\cite{zhai2023sigmoid}, CLIP~\cite{radford2021learning}, and SD-VAE~\cite{van2017neural}}
    \centering
  \resizebox{0.8\linewidth}{!}{
    \begin{tabular}{l|cc|c}
    \toprule
\multirow{2}{*}{Methods} & \multicolumn{2}{c|}{Adroit} & \\
 & Pen & Relocate  & Mean\\
    \midrule
    \midrule
w/o vision encoder &  76.0 \footnotesize{$\pm$4.0} & 33.3 \footnotesize{$\pm$2.3} & 54.7 \\
+ SigLIP~\cite{zhai2023sigmoid} &  77.3 \footnotesize{$\pm$2.3} & \phantom{0}33.3 \footnotesize{$\pm$10.1} & 55.3 \\
+ CLIP~\cite{radford2021learning} &  81.3 \footnotesize{$\pm$4.6} & \phantom{0}\textbf{45.3} \footnotesize{$\pm$14.6} & 63.3 \\
+ SD-VAE~\cite{van2017neural} & 81.3 \footnotesize{$\pm$6.1} & 41.3 \footnotesize{$\pm$2.3} & 61.3 \\
\midrule

\textbf{\textbf{ORCA} (Ours)} & \textbf{86.7} \footnotesize{$\pm$2.3} & \text{44.0} \footnotesize{$\pm$4.0} & \textbf{65.3}  \\
  \bottomrule

  \end{tabular}}
    \label{tab:ablation}
    \end{table}

In Table ~\ref{tab:ablation}, we provide ablation on the choice of the additional visual encoder $\mathcal{E}_V$, which is utilized to obtain the visual prompts $p_v$. In this regard, we replace the DINOv2~\cite{oquab2023dinov2} encoder with SigLIP~\cite{zhai2023sigmoid} and CLIP~\cite{radford2021learning}, as well as SD-VAE~\cite{van2017neural} which is used for obtaining the latents in Stable Diffusion 1.5. 
As shown in Table~\ref{tab:ablation}, all variants show noticeable gains over the baseline (i.e., w/o vision encoder), while the choice of encoder shows some variance in terms of performance. Especially, the results from SD-VAE shows that we could achieve competitive results even without an external vision encoder. This confirms that the usage of visual prompt itself provides a substantial benefit and plays a core role in its overall effectiveness. 

\subsection{Ablation on diffusion timesteps} \label{sec:timesteps}
\begin{table}[h]
\vspace{-10pt}
  \caption{\textbf{Ablation study on timestep selection}. To ablate the choice for timestep $t$, we provide results with $t=100$ and $t=200$. The performance of imitation learning agents on DeepMind Control~\citep{tassa2018deepmind} is reported. We report the normalized score averaged over three seeds with its standard deviation.}
  
  \label{tab:timesteps}
  \centering
  \resizebox{\linewidth}{!}{
  \begin{tabular}{l|ccccc|c}
    \toprule
 & \multicolumn{5}{c|}{DeepMind Control} &\\
Timestep &  Walker-stand & Walker-walk & Reacher-easy & Cheetah-run & Finger-spin & Mean\\
    \midrule
    \midrule

200 & \textbf{92.2} \footnotesize{$\pm$ 1.6} & \textbf{78.6} \footnotesize{$\pm$ 2.2} & \text{85.4} \footnotesize{$\pm$ 8.3} & 24.7 \footnotesize{$\pm$ 4.5} & \text{66.5} \footnotesize{$\pm$ 3.2} & \text{69.4} \\

100 & 88.3 \footnotesize{$\pm$ 4.7} & 72.6 \footnotesize{$\pm$ 4.3} & \text{79.4} \footnotesize{$\pm$ 6.8} & \text{36.1} \footnotesize{$\pm$ 6.2} & 66.2 \footnotesize{$\pm$ 3.5} & 68.5 \\

\midrule

\textbf{0} (Default) & \text{89.1} \footnotesize{$\pm$ 1.8} & 76.9 \footnotesize{$\pm$ 4.0}  & \textbf{87.6} \footnotesize{$\pm$ 2.9}  & \textbf{50.0} \footnotesize{$\pm$ 8.4}  & \textbf{68.0} \footnotesize{$\pm$ 1.0} & \textbf{74.3} \\
\bottomrule
  \end{tabular}}
\end{table}

To ablate the effects of the diffusion timestep $t$, in Table ~\ref{tab:timesteps}, we additionally provide results with $t=100$ and $t=200$. Although some tasks (\textit{e.g.} \textit{Reacher-easy}) benefit from $t=100$ or $t=200$, performance on other tasks such as \textit{Cheetah-run} degrades significantly, lowering the overall score. Therefore, we choose $t=0$, which achieves the best overall performance.

\subsection{Comparison with stronger pre-trained visual representations} \label{sec:encoders}

\begin{table}[h]
\caption{\textbf{Additional comparison with stronger pre-trained visual representations}. We further provide comparison with recent state-of-the-art pre-trained visual representations, including DINOv2~\cite{oquab2023dinov2}, SigLIP~\cite{zhai2023sigmoid}, and Theia~\cite{shang2024theia}. We report the success rates (\%) averaged over three seeds with their standard deviation.}
  \centering

  \resizebox{\linewidth}{!}{
  \begin{tabular}{l|ccccc|c}
    \toprule
 \multirow{2}{*}{Method}& \multicolumn{5}{c|}{DeepMind Control} &\\
 &  Walker-stand & Walker-walk & Reacher-easy & Cheetah-run & Finger-spin & Mean\\
    \midrule
    \midrule

DINOv2~\cite{oquab2023dinov2} & 87.6 {$\pm$5.2} & 61.4 {$\pm$9.3} & \text{22.5} {$\pm$3.2} & \text{27.1} {$\pm$4.2} & \phantom{0}64.1 {$\pm$10.8} & 52.5 \\
SigLIP~\cite{zhai2023sigmoid} & 81.4 {$\pm$3.6} & 50.3 {$\pm$4.4} & \text{87.1} {$\pm$5.5} & \text{18.6} {$\pm$2.6} & 66.6 {$\pm$2.4} & 60.8 \\
Theia~\cite{shang2024theia} & 85.3 {$\pm$2.8} & 65.0 {$\pm$2.9} & \phantom{0}\text{61.4} {$\pm$13.4} & \text{43.8} {$\pm$9.8} & 67.8 {$\pm$1.0} & 64.6 \\

\midrule

\textbf{\textbf{ORCA} (Ours)} & \textbf{89.1} {$\pm$1.8} & \textbf{76.9} {$\pm$4.0}  & \textbf{87.6} {$\pm$2.9}  & \textbf{50.0} {$\pm$8.4}  & \textbf{68.0} {$\pm$1.0} & \textbf{74.3} \\
\bottomrule
  \end{tabular}}
    \label{tab:rep}
\end{table}

To study whether recent state-of-the-art pre-trained visual representations could overcome its limitations from task-agnostic nature, we compare ORCA with DINOv2~\cite{oquab2023dinov2} and SigLIP~\cite{zhai2023sigmoid}, while also including Theia (Shang et al., 2024), a distilled vision foundation model built for manipulation tasks in Table~\ref{tab:rep}. Notably, these pre-trained encoders still exhibit underwhelming performance in several tasks (\textit{e.g.} Reacher-easy for DINOv2, Cheetah-run for SigLIP), highlighting the inherent limitations of task-agnostic representations in complex control settings. Considering that SigLIP is widely adopted in recent Vision-Language-Action (VLA) models, these results suggest that incorporating ORCA within VLA models can be a promising direction for future research.

\subsection{Efficiency comparison} \label{sec:efficiency}

\begin{table}[h]
  \caption{\textbf{Efficiency comparison}. We report the total number of parameters (\#Params), the number of learnable parameters (\#Learnable) and latency for VC-1, SCR, and ours.}
  \label{tab:efficiency}
  \centering
  \resizebox{0.7\linewidth}{!}{
  \begin{tabular}{l|c|c|c}
    \toprule
    Methods & \#Params & \#Learnable & Time \\
    \midrule
    \midrule
    VC-1~\cite{majumdar2023we} & 303.3M & 0 & 11ms\\
    SCR~\cite{gupta2024pre}& 382.9M & 0 & 26ms \\
    Ours& 480.1M & 10.6M & 48ms\\
    \bottomrule
  \end{tabular}}
\end{table}

In Table~\ref{tab:efficiency}, we report the number of parameters, number of learnable parameters, and latency for each modules for VC-1~\citep{majumdar2023we}, SCR~\citep{gupta2024pre} and our proposed method. For VC-1 and SCR, we use ViT-L/16, which was also used for comparison in the main paper. Notably, the layer selection allows us to drop the “up” blocks, which removes around 500M parameters from the denoising U-Net. This allows the U-Net to have similar parameter count to VC-1 encoder, which is used in various robotic manipulation tasks. Furthermore, most of the parameters added to our method are the frozen parameters from DINOv2, and the learnable parameters consist mostly of the additional projection layers for the visual prompts.

\subsection{Analysis on the null condition} \label{sec:null}
\begin{figure}[h]
  \centering
  \includegraphics[width=\linewidth]{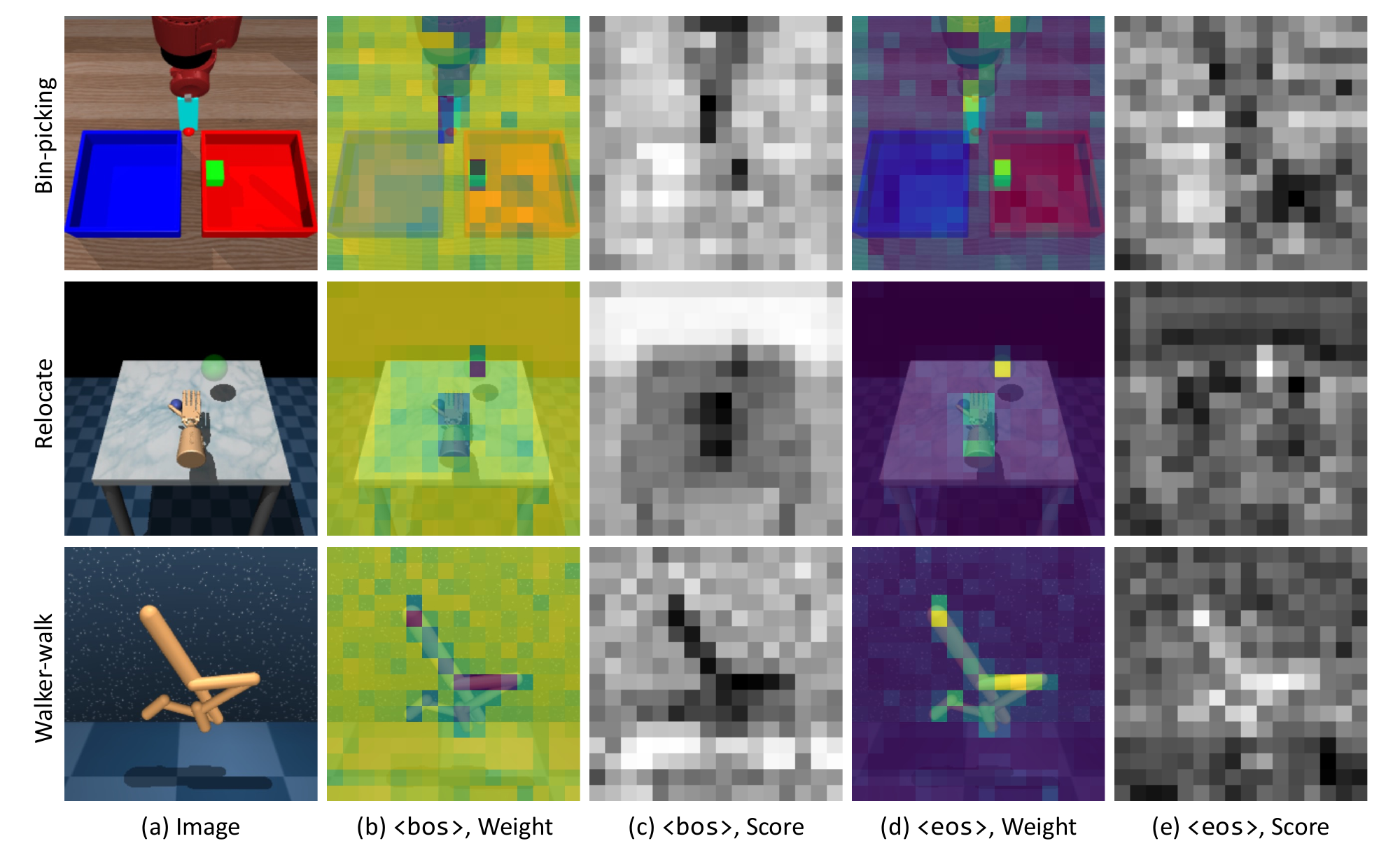}
  \caption{\textbf{Visualization of normalized attention weights and raw attention scores for \texttt{<bos>} and \texttt{<eos>} tokens.} We compare the visualization of the normalized attention weights obtained after the softmax operation and the raw attention scores obtained before the softmax operation from the cross-attention layers to further analyze the properties of \texttt{<bos>} and \texttt{<eos>} tokens.}
  \label{fig:supp}
\end{figure}

Figure~\ref{fig:supp} illustrates the attention behavior of \texttt{<bos>} and \texttt{<eos>} tokens by visualizing their normalized cross-attention maps (b,d) and raw attention scores (c,e). We observe that the \texttt{<bos>} token consistently attends to background regions, whereas the \texttt{<eos>} is less reliable at focusing on salient objects (\textit{e.g.} the robot hand in \textit{Relocate}). We attribute the background affinity of \texttt{<bos>} to the typical structure of text prompts, which primarily describe foreground subjects. Moreover, since Stable Diffusion employs a causal text encoder for both conditional prompts and the null condition $\varnothing$ in unconditional generation, this background-attending behavior is also transferred to unconditional scenarios.

\subsection{Further discussions} \label{sec:supp_dis}
It is also worth considering Vision-Language-Action models (VLAs)~\cite{o2024open, kim2024openvla}, which are trained on large-scale robotic data. However, adapting these billion-scale models to new robot embodiments often necessitates computationally expensive fine-tuning~\cite{kim2024openvla}. While this cost is arguably mitigated by their generalization capabilities, such as zero-shot instruction following from language descriptions, we focus on efficient, vision-based specialist agents for robotic control tasks rather than generalist approach. Consequently, ensuring these specialist agents are task-adaptive remains a critical research objective. Furthermore, it is important to note that conditional diffusion models leverage general image-text data~\citep{schuhmann2022laion}, which is significantly easier to collect and scale than the robot interaction data required to train VLAs~\citep{o2024open}.

\section{Further Implementation Details}
\label{sec:supp_impl}

\subsection{Full description of the text conditions}\label{sec:prompt}
In Table~\ref{tab:prompt}, we provide the full descriptions used for Text (Simple) and Text (Caption), which are generated by Gemini 2.5 Pro~\citep{comanici2025gemini}. For CoOp~\citep{zhou2022learning}, we use 4 learnable prefix tokens, such as ``[$V^*$][$V^*$][$V^*$][$V^*$] bin picking" for \textit{Bin-picking}. For TADP, we add a style prefix ``in a [$S^*$] style", which results in ``The Sawyer robot arm must carefully pick a specific target object out of the cluttered red bin and place it into the empty blue bin in a [$S^*$] style." for \textit{Bin-picking}.

\begin{table*}[t]
\caption{\textbf{Full text descriptions used in baselines}.}
  \label{tab:prompt}
  \centering
  \resizebox{\linewidth}{!}{
  \begin{tabular}{l|c|c}
    \toprule
Task & Method & Text\\
    \midrule
    \midrule
\multirow{2}{*}{Assembly} & Text (Simple) & ``assembly"\\
 & Text (Caption) & ``The Sawyer robot arm must pick up the green block and precisely insert it into the center of the silver ring to complete the assembly."\\

 \multirow{2}{*}{Bin} & Text (Simple) & ``bin picking"\\
 & Text (Caption) & ``The Sawyer robot arm must carefully pick a specific target object out of the cluttered red bin and place it into the empty blue bin."\\

 \multirow{2}{*}{Button} & Text (Simple) & ``button press"\\
 & Text (Caption) & ``The Sawyer robot arm must reach out and accurately press the red button on top of the yellow control box."\\

 \multirow{2}{*}{Drawer} & Text (Simple) & ``drawer open"\\
 & Text (Caption) & ``The Sawyer robot arm must grasp the white handle and pull open the light green drawer."\\

 \multirow{2}{*}{Hammer} & Text (Simple) & ``hammer"\\
 & Text (Caption) & ``The Sawyer robot arm must pick up the red hammer and use it to strike the nail, driving it completely into the wooden block."\\

 \midrule

\multirow{2}{*}{Pen} & Text (Simple) & ``pen"\\
 & Text (Caption) & ``A dexterous robotic hand must twirl a blue pen within its grasp to match the final orientation shown by the green target pen."\\

\multirow{2}{*}{Relocate} & Text (Simple) & ``relocate"\\
 & Text (Caption) & ``A dexterous robotic hand is tasked with picking up the small blue ball and moving it to the location of the green target sphere."\\

\midrule

\multirow{2}{*}{Cheetah-run} & Text (Simple) & ``cheetah run"\\
 & Text (Caption) & ``A minimalist orange robot, shaped like a cheetah, runs across a reflective floor in a simulated environment."\\

\multirow{2}{*}{Walker-walk} & Text (Simple) & ``walker walk"\\
 & Text (Caption) & ``A minimalist, orange bipedal robot takes a step across a reflective floor in a simulated environment."\\

\multirow{2}{*}{Walker-stand} & Text (Simple) & ``walker stand"\\
 & Text (Caption) & ``A minimalist, orange bipedal robot stands upright on a reflective floor in a simulated environment."\\
 
 \multirow{2}{*}{Finger-spin} & Text (Simple) & ``finger spin"\\
 & Text (Caption) & ``A simple robotic finger strikes a floating, hot dog-shaped object to make it spin against a starry background."\\
 
 \multirow{2}{*}{Reacher} & Text (Simple) & ``reacher"\\
 & Text (Caption) & ``A simple robotic arm reaches for a red target ball on a checkered blue surface."\\

\bottomrule
  \end{tabular}}
\end{table*}

\subsection{Details of the baselines} \label{sec:baselines}
\noindent\textbf{CLIP}~\citep{radford2021learning} is a vision-language model pre-trained on large-scale image-text pairs through contrastive learning. CLIP has been widely used in various tasks, including navigation and manipulation tasks~\citep{shridhar2022cliport, khandelwal2022simple}. 

\noindent\textbf{VC-1}~\citep{majumdar2023we} is a foundation model for various robotics tasks, spanning from manipulation to locomotion and navigation tasks. VC-1 trains with MAE objective on egocentric videos, as well as additional data including navigation and manipulation datasets.

\noindent\textbf{SCR}~\citep{gupta2024pre} employs Stable Diffusion for various navigation and manipulation tasks. We consider SCR as a baseline using the null condition $\varnothing$, which is implemented as an empty string. 

\noindent\textbf{Text(Simple/Caption)} is a task-adaptive baseline using text conditions, where Text (Simple) directly uses the task names as the condition, whereas Text (Caption) leverages descriptions generated from Gemini 2.5~\citep{comanici2025gemini}. Full text used for each tasks are presented in the appendix.

\noindent\textbf{CoOp}~\citep{zhou2022learning} extends on Text$_{\text{simple}}$ by implementing learnable prefix tokens $V^*$. CoOp originally prompts CLIP with the format ``[$V^*$] \textit{classname}" for image classification, which in our case, the task names used in Text$_{\text{simple}}$ are used as classnames.

\noindent\textbf{TADP}~\citep{kondapaneni2024text} extends on Text$_{\text{caption}}$, by adding a special token $S^*$ that encapsulates the visual style information optimized through Textual Inversion~\citep{gal2022image}. Since the visual information is optimized into a single token $S^*$, we can consider TADP as a baseline with global visual information, and not in a frame-wise manner.

\subsection{Implementation details of the compression layer}\label{sec:supp_compression}
\vspace{-5pt}
\begin{algorithm}[h]
\SetAlgoLined
    \PyCode{class CompressionLayer(nn.Module):} \\
    \Indp   
        \PyCode{def \_\_init\_\_(self, hidden\_dim, compress\_dim):} \\
        \Indp
            \PyCode{self.layers = nn.Sequential(} \\
            \Indp
                \PyCode{nn.Conv2d(hidden\_dim, compress\_dim, kernel\_size=3, padding=1), }\\
                \PyCode{nn.BatchNorm2d(hidden\_dim),}\\
                \PyCode{nn.ReLU(inplace=True),}\\
                \PyCode{nn.Flatten()}\\
                \PyCode{)}\\
            \Indm
        \Indm
    \Indm 
    \Indp   
        \PyCode{def forward(self, x):} \\
        \Indp
            \PyCode{return self.layers(x)} \\
        \Indm
    \Indm
\caption{PyTorch-style pseudocode for the compression layer}
\label{algo:your-algo}
\end{algorithm}

To provide further details of the compression layer~\citep{yadav2023ovrl}, we provide a PyTorch-style pseudo-code of the compression layer in Alg.~\ref{algo:your-algo}. We follow previous works~\citep{yadav2023ovrl, gupta2024pre} for implementing a simple convolutional layer for the compression layer to obtain 1D state representations from 2D features. For all methods, \texttt{compress\_dim} was set to 48. Note that the compression layer was also used for compared baselines including CLIP~\citep{cherti2023reproducible} and VC-1~\citep{majumdar2023we}, which have been shown to yield better performance than using \texttt{<CLS>} tokens~\citep{gupta2024pre}.

\section{Limitations}\label{sec:limitations}
As previously discussed, we base our exploration on pre-trained Stable Diffusion v1.5, which is a U-Net-based diffusion model. Consequently, our findings may not directlyt apply to diffusion models with different architectures, such as DiT~\cite{peebles2023scalable, esser2024scaling} models or video diffusion models~\cite{blattmann2023stable}, which we leave to be further explored in the future.

\section{Qualitative visualization on robotic control tasks}\label{sec:qual}

\begin{figure*}[h]
  \centering
  \includegraphics[width=1.0\linewidth]{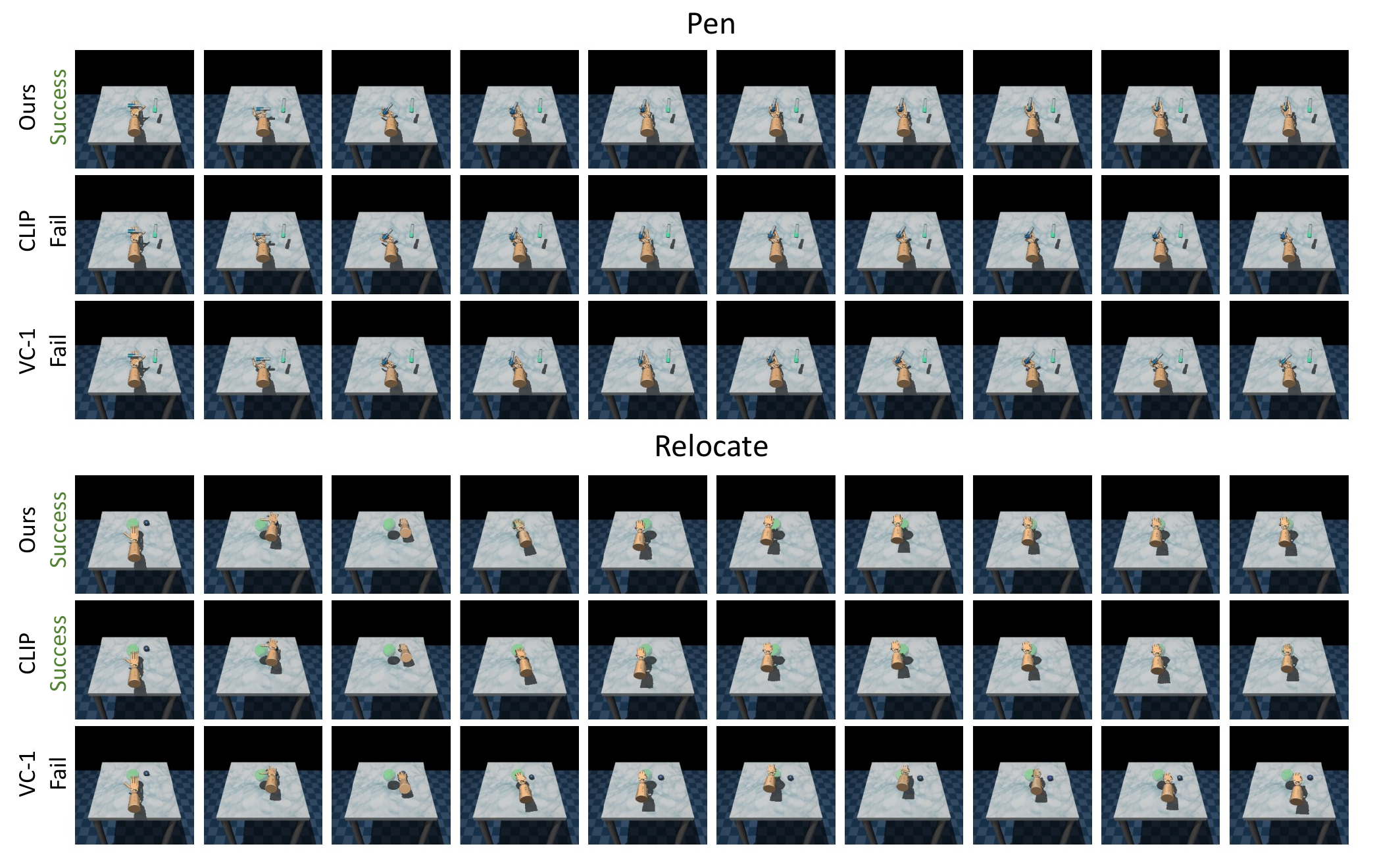}
  \caption{\textbf{Visualization of agents performing downstream tasks in Adroit~\citep{adroit}.} We provide visual comparison of our method to CLIP~\citep{cherti2023reproducible}, and VC-1~\citep{majumdar2023we} for two tasks from Adroit. We additionally report whether the task has succeeded or failed for each episode.}
  \label{fig:video_adroit}
  \vspace{-15pt}
\end{figure*}

\begin{figure*}[h]
  \centering
  \includegraphics[width=0.8\linewidth]{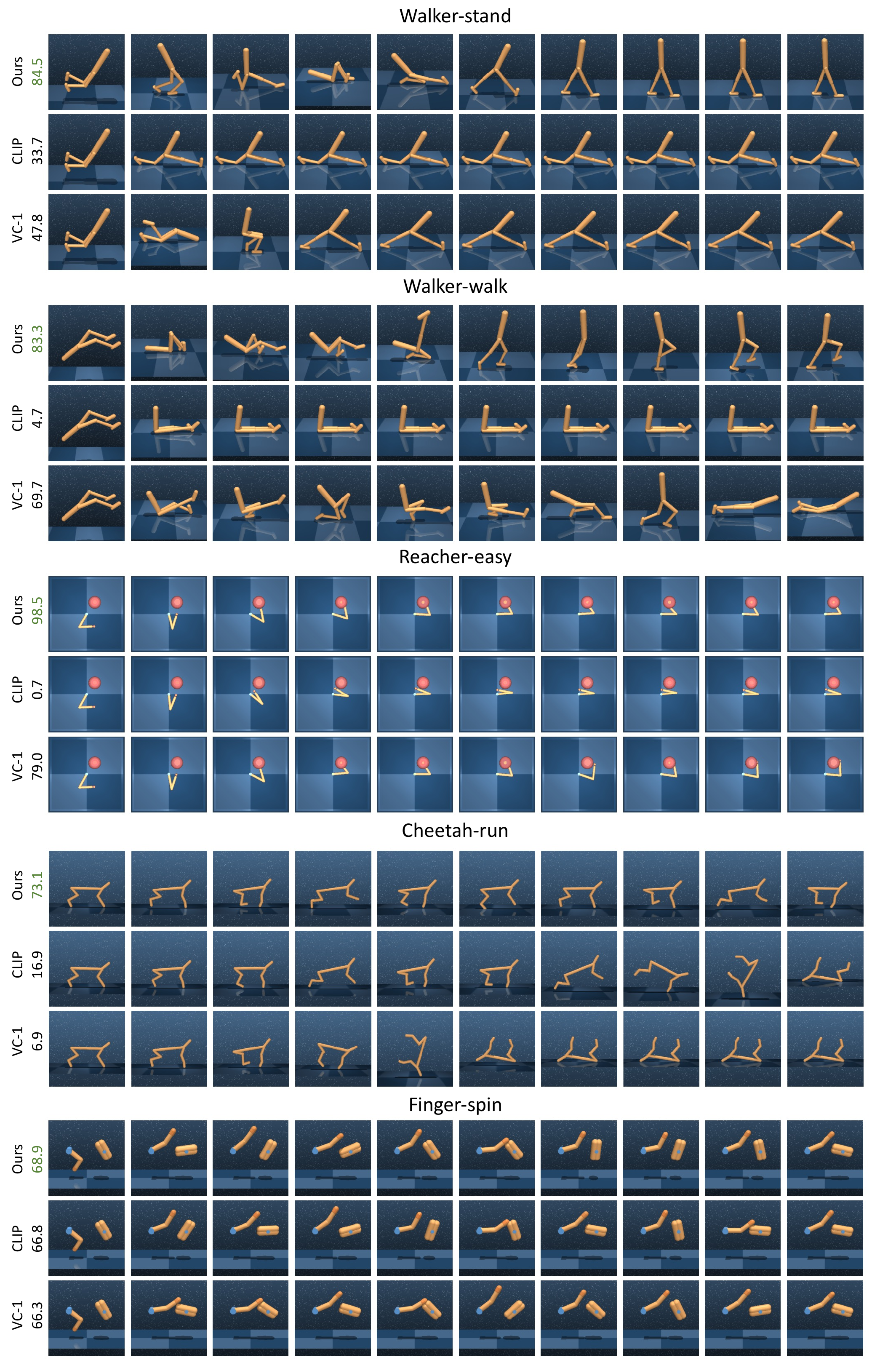}
  \caption{\textbf{Visualization of agents performing downstream tasks in DMC~\citep{tassa2018deepmind}.} We provide a visual comparison of our method to CLIP~\citep{cherti2023reproducible}, and VC-1~\citep{majumdar2023we} for five tasks in DMC. We additionally report the normalized score for each episode.}
  \label{fig:video_dmc}
  \vspace{-15pt}
\end{figure*}

\begin{figure*}[h]
  \centering
  \includegraphics[width=0.8\linewidth]{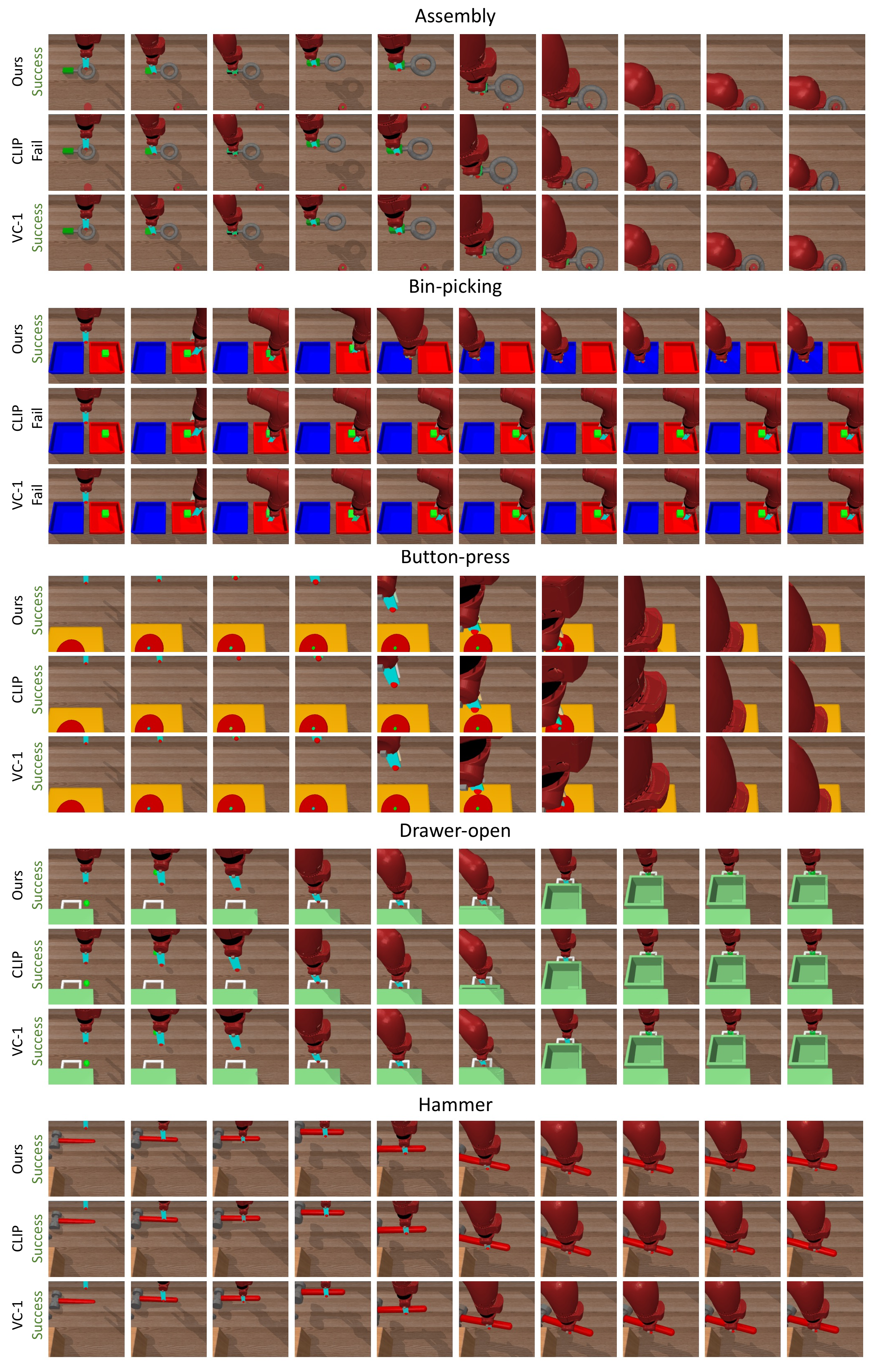}
  \caption{\textbf{Visualization of agents performing downstream tasks in MetaWorld~\citep{metaworld}.} We provide visual comparison of our method to CLIP~\citep{cherti2023reproducible}, and VC-1~\citep{majumdar2023we} for five tasks in MetaWorld. We additionally report whether the task has succeeded or failed for each episode.}
  \label{fig:video_meta}
  \vspace{-15pt}
\end{figure*}

We provide frame-wise comparison of our method, CLIP~\citep{cherti2023reproducible}, and VC-1~\citep{majumdar2023we} for tasks from DMC~\citep{tassa2018deepmind} in Fig.~\ref{fig:video_dmc}, MetaWorld~\citep{metaworld} in Fig.~\ref{fig:video_meta}, and Adroit~\citep{adroit} in Fig.~\ref{fig:video_adroit}. For each task, we report the normalized score for DMC and whether the task has succeeded or failed for MetaWorld and Adroit. 

\clearpage

{
    \small
    \bibliographystyle{ieeenat_fullname}
    \clearpage
    \bibliography{main}
}
\end{document}